\documentclass[final,12pt]{elsarticle}

\makeatletter
\def\ps@pprintTitle{}
\makeatother

\usepackage{microtype}

\makeatletter
\renewcommand{\subsection}{\@startsection{subsection}{2}{\z@}{-12pt}{6pt}{\normalsize\bfseries}}
\renewcommand{\subsubsection}{\@startsection{subsubsection}{3}{\z@}{-12pt}{6pt}{\normalsize\bfseries}}
\makeatother

\usepackage{amsmath, amssymb, amsthm}
\usepackage{bm}

\usepackage{graphicx}
\usepackage{subcaption}
\usepackage{booktabs}
\usepackage{multirow}
\usepackage{float}

\usepackage{algorithm}
\usepackage{algpseudocode}

\usepackage{hyperref}
\usepackage[capitalize]{cleveref}

\usepackage{xcolor}

\newcommand{\adrasj}{ADRAS-J}
\newcommand{\colmap}{COLMAP}
\newcommand{\neuralangelo}{Neuralangelo}
\newcommand{\samthree}{SAM3}
\newcommand{\ppisp}{PPISP}
\newcommand{\sfm}{SfM}

\newcommand{\rso}{RSO}

\title{From Images2Mesh: A 3D Surface Reconstruction Pipeline for Non-Cooperative Space Objects\tnoteref{t1}}

\tnotetext[t1]{The STS-119 and ADRAS-J footage used in this work are publicly available. This work was conducted independently and is not affiliated with or endorsed by NASA, JAXA, or Astroscale Japan.}

\author[a]{Bala Prenith Reddy Gopu}
\author[a]{Patrick Quinn}
\author[a]{George M. Nehma}
\author[a]{Madhur Tiwari}
\author[b]{Matt Ueckermann}
\author[b]{David Hinckley}
\author[b]{Christopher McKenna}

\affiliation[a]{
    organization={Department of Aerospace, Physics and Space Sciences, Florida Institute of Technology},
    addressline={150 W University Blvd},
    city={Melbourne},
    postcode={32901},
    state={FL},
    country={USA}
}

\affiliation[b]{
    organization={Creare LLC},
    city={Hanover},
    postcode={03755},
    state={NH},
    country={USA}
}

\date{}

\begin{document}

\begin{abstract}
On-orbit inspection imagery is crucial as it enables characterization of non-cooperative resident space objects, providing the geometry and structural condition essential for active debris removal and on-orbit servicing mission planning. However, most existing neural implicit surface reconstruction methods have been confined to synthetic or hardware-in-the-loop data with known camera poses and controlled illumination. In this work, we present a pipeline for neural implicit surface reconstruction of non-cooperative space objects from monocular inspection imagery. We demonstrate it on publicly released ISS inspection footage from the STS-119 mission and publicly released on-orbit inspection footage of an H-IIA rocket upper stage. We find that segmentation-based background removal is essential for successful camera pose estimation from real on-orbit footage, where background variation between frames caused direct processing to fail entirely. We further incorporate photometric correction of per-frame exposure variations and analyze its behavior across datasets, finding that performance in shadowed regions varies with the illumination characteristics of the input footage. \footnote{Video results will be made available at \url{https://www.youtube.com/@theautonomylab}}
\end{abstract}

\maketitle

\section{Introduction}
\label{sec:intro}

The growing population of uncontrolled resident space objects in low Earth orbit poses an escalating risk to operational satellites and future space missions. Characterizing these objects, understanding their geometry, structural condition, and rotational dynamics, is a critical prerequisite for active debris removal and on-orbit servicing missions, yet remains an open challenge. In July 2024, the \adrasj{} mission marked a historic milestone by becoming the first spacecraft to perform rendezvous and proximity operations with a large defunct rocket body, conducting a fly-around maneuver at a close range of approximately 50 meters to capture inspection imagery \cite{jaxa2024crd2, astroscale2024flyaround}. The publicly released inspection footage from this mission, alongside publicly released footage from NASA's STS-119 mission, in which Space Shuttle Discovery performed a fly-around of the ISS in 2009, represent examples of the real on-orbit imagery that motivates the need for reconstruction pipelines capable of operating directly on such data without ground truth camera poses or a reference CAD model.

While autonomous detection and tracking of \rso{}s represent a critical first step toward space situational awareness \cite{zuehlke2022autonomous}, knowing where an object is tells us little about what it looks like, how it is oriented, or what condition it is in. Safe and successful capture of a non-cooperative \rso{} requires a detailed understanding of its 3D geometry and structural condition, information that cannot be obtained from detection or tracking alone. Without this knowledge, mission planners cannot assess the risks involved in approaching a tumbling, potentially fragile object, plan a safe capture trajectory, or simulate removal operations in advance. 3D reconstruction from inspection imagery offers a promising path toward this level of characterization, enabling mission planners to assess the target before committing to a capture attempt and supporting the generation of synthetic datasets for autonomous inspection research \cite{quinn2026simulation}.

Prior work on neural implicit surface reconstruction of \rso{}s has been confined to controlled simulation environments and hardware-in-the-loop testbeds, where camera poses and lighting conditions are known by design. 3DGS has been applied to real on-orbit inspection footage from the \adrasj{} mission, motivated by its computational efficiency and suitability for onboard deployment \cite{hopkins2025scoutspace, issitt2025optimal}. However, no prior work has produced an explicit high-fidelity surface mesh from real on-orbit inspection imagery without ground truth camera poses or a reference CAD model. 

In this work, we present a five-stage pipeline for neural implicit surface reconstruction of non-cooperative space objects from monocular inspection imagery. Our pipeline integrates temporal frame extraction, background removal via \samthree{}, structure from motion with \colmap{}, neural implicit surface reconstruction via \neuralangelo{}, and photometric post-processing with \ppisp{}. We demonstrate that background removal is a critical prerequisite for successful camera registration on real on-orbit footage, where direct processing of raw frames failed entirely.

The main contributions of this work are as follows:
\begin{itemize}
    \item We present a five-stage pipeline for neural implicit surface reconstruction of non-cooperative space objects from real on-orbit inspection imagery, requiring no ground truth camera poses or reference CAD model.
    \item We integrate photometric correction of per-frame exposure variations into the pipeline and analyze its behavior across datasets, demonstrating qualitative improvements on real on-orbit footage of a simple object while identifying limitations on more complex structures.
    \item The meshes produced by the pipeline can directly support future active debris removal mission planning, enabling debris characterization, structural assessment, and simulation-based capture strategy development, as well as generation of synthetic datasets for autonomous inspection research.
\end{itemize}

\section{Related Work}
\label{sec:related}

\subsection{Non-cooperative Space Object Pose and Shape Estimation}
Most existing work on \rso{} characterization has been conducted in simulation, using synthetic datasets where ground truth poses and lighting conditions are fully controlled. Lightweight convolutional neural networks have been proposed for real-time detection and localization of spacecraft features such as solar panels, antennas, and thrusters, trained on web-scraped satellite imagery consisting primarily of synthetic renderings and evaluated on hardware-in-the-loop testbeds \cite{mahendrakar2023spaceyolo, mahendrakar2024unknown}. While these methods are well-suited for rapid feature detection, they do not recover the 3D geometry of the \rso{}, which is necessary for detailed inspection and mission planning.

A CNN-based approach was proposed to jointly estimate the pose and 3D structure of an unknown \rso{} from a single 2D image, representing the shape as an assembly of superquadric primitives that can describe a range of geometric forms using only a small number of parameters \cite{park2024rapid}. While this enables rapid, single-shot characterization without requiring multiple views, the recovered shapes are inherently coarse and capture only the high-level structure of the target. A follow-on work addressed body-axis ambiguities that arise during monocular shape and pose estimation, improving the reliability of the recovered pose \cite{bates2025removing}. The coarse superquadric shapes produced by these methods were further leveraged as geometric priors to initialize 3D Gaussian Splatting, enabling faster convergence toward a higher-fidelity reconstruction 
of the \rso{} \cite{hucfast}.

3D Gaussian Splatting has been applied to characterize the geometry of unknown \rso{}s from imagery, demonstrating the ability to learn high-quality 3D representations on hardware-in-the-loop satellite mockups within the computational constraints of spaceflight hardware \cite{nguyen2024characterizing}. Building on this, optimal inspection orbits for 3DGS-based \rso{} characterization were investigated using a Blender-based digital twin of the orbital environment, finding that V-bar maneuvers consistently yield the best reconstruction quality across satellite geometries and altitudes \cite{issitt2025optimal}. 
However, these works rely on synthetic or hardware-in-the-loop data with controlled illumination and do not recover an explicit 3D surface mesh of the \rso{}.

\subsection{Neural Implicit Surface Reconstruction for Space Objects}
NeRF \cite{mildenhall2021nerf} and 3DGS \cite{kerbl3Dgaussians} represent two complementary approaches to learning 3D scene representations from 2D images, and both have been explored for the reconstruction of spacecraft in the space environment. A comparison between NeRF and Generative Radiance Fields (GRAF) on synthetic spacecraft imagery demonstrated the feasibility of neural representations for spacecraft novel view synthesis and 3D shape extraction, with NeRF requiring known camera poses and GRAF operating without pose information \cite{mergy2021vision}. Instant-NGP and D-NeRF were applied to RSO reconstruction using hardware-in-the-loop imagery of a satellite mockup, finding that Instant-NGP achieved competitive reconstruction quality at significantly reduced computational cost, while D-NeRF offered no advantage over static variants for this application \cite{caruso20233d}. Instant-NGP was also applied to hardware-in-the-loop satellite imagery for 3D mesh reconstruction, outperforming a multi-view stereo baseline in both reconstruction fidelity and training time \cite{huber2024high}. \neuralangelo{} \cite{li2023neuralangelo} introduced higher-fidelity neural implicit surface reconstruction compared to standard NeRF variants, and was applied to synthetic RSO reconstruction using ground truth camera poses from a physics-based simulator, demonstrating that accurate pose estimation is a prerequisite for high-fidelity \neuralangelo{} reconstruction under controlled illumination conditions \cite{gopu2026dynamic}. A shared limitation across NeRF, \neuralangelo{}, and 3DGS is the assumption of photometric consistency across views, which presents a significant challenge in the space environment where illumination varies continuously and spacecraft structures exhibit strong specular reflectance. 3DGS was extended to model dynamic illumination in the space environment by incorporating Sun vector knowledge and a shadow splatting mechanism to capture global shadowing and self-occlusion, though the approach also assumes known camera poses and initializes Gaussians from a ground-truth 3D model surface \cite{park2025improved}. \ppisp{} offers a more general approach to photometric correction in multi-view reconstruction by learning per-frame photometric variations directly from the imagery, without requiring prior knowledge of the illumination geometry or Sun vector \cite{deutsch2026ppispphysicallyplausiblecompensationcontrol}. While prior applications of neural implicit surface reconstruction to spacecraft rely exclusively on synthetic or hardware-in-the-loop data with known camera poses, 3DGS has been applied to real on-orbit inspection imagery \cite{hopkins2025scoutspace, issitt2025optimal}.

\section{Methodology}
\label{sec:Methodology}

\subsection{Pipeline Overview}

\begin{figure*}[t]
    \centering
    \includegraphics[width=\textwidth]{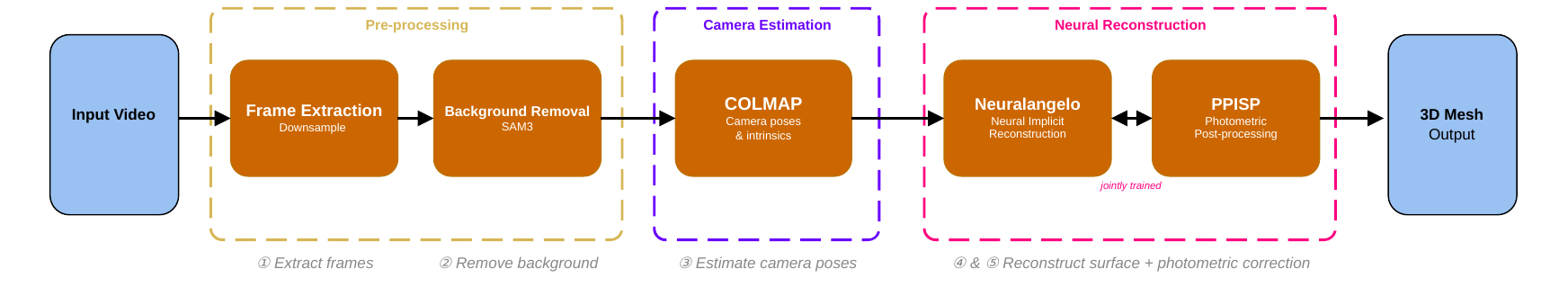}
    \caption{Overview of our pipeline comprising temporal frame extraction, background removal using SAM3, structure from motion with COLMAP, neural implicit reconstruction via Neuralangelo, and photometric post-processing with PPISP.}
    \label{fig:pipeline}
\end{figure*}

In this section, we present an overview of our pipeline. Our pipeline consists of five sequential stages comprising temporal frame extraction, background removal using \samthree{}, structure from motion with \colmap{}, neural implicit reconstruction via \neuralangelo{}, and photometric post-processing with \ppisp{}. First, we extract the frames from the input inspection video using a dataset specific downsampling factor. We then pre-process the extracted frames using a state-of-the-art segmentation algorithm, \samthree{}, to remove the background. After pre-processing, we run \colmap{} to extract the camera intrinsic matrix and pose of the camera for each frame. Once we have the camera poses, we train a neural implicit surface reconstruction algorithm, \neuralangelo{}, to generate a 3D mesh from the learned implicit representation. To address photometric variations inherent in on-orbit inspection imagery, we integrate a post-processing algorithm \ppisp{} that learns these variations alongside \neuralangelo{} and applies photometric corrections. The full pipeline is illustrated in Figure \ref{fig:pipeline}.

\subsection{Temporal Frame Extraction}
\label{sec:frames}

We extract frames from the input inspection video using FFmpeg with a dataset specific downsampling factor, specifically selecting every $n$-th frame from the video. The specific downsampling factor and resulting frame count for each dataset are described in Section~\ref{sec:dataset}.

\subsection{Background Removal via \samthree{}}
\label{sec:segmentation}

Upon running \colmap{} \cite{schoenberger2016sfm, schoenberger2016mvs} directly on the raw frames, we observed that only a few frames were successfully registered across multiple experiments in which we varied the feature matching strategy (sequential vs exhaustive) and the overlap parameter. Consequently, we suspected that the inconsistent background, wherein we had a black background in some frames and Earth as the background in others, could be contributing to this poor registration performance. Therefore, we explored segmentation methods to isolate the foreground from the background. We initially experimented with methods such as threshold-based and color-based masking to isolate the foreground, but the presence of Earth in the background introduced complexity beyond what fixed thresholds and color ranges could handle; furthermore, they required significant manual tuning for each image. Given these limitations, we adopted an automatic segmentation method, \samthree{} \cite{carion2025sam3segmentconcepts}. \samthree{} is a segmentation model that extends the Segment Anything Model by introducing text-based prompting, enabling the model to understand and segment the object of interest without requiring manual point or box annotations.

\begin{figure}
    \centering

    \begin{subfigure}[b]{0.32\textwidth}
        \includegraphics[width=\textwidth]{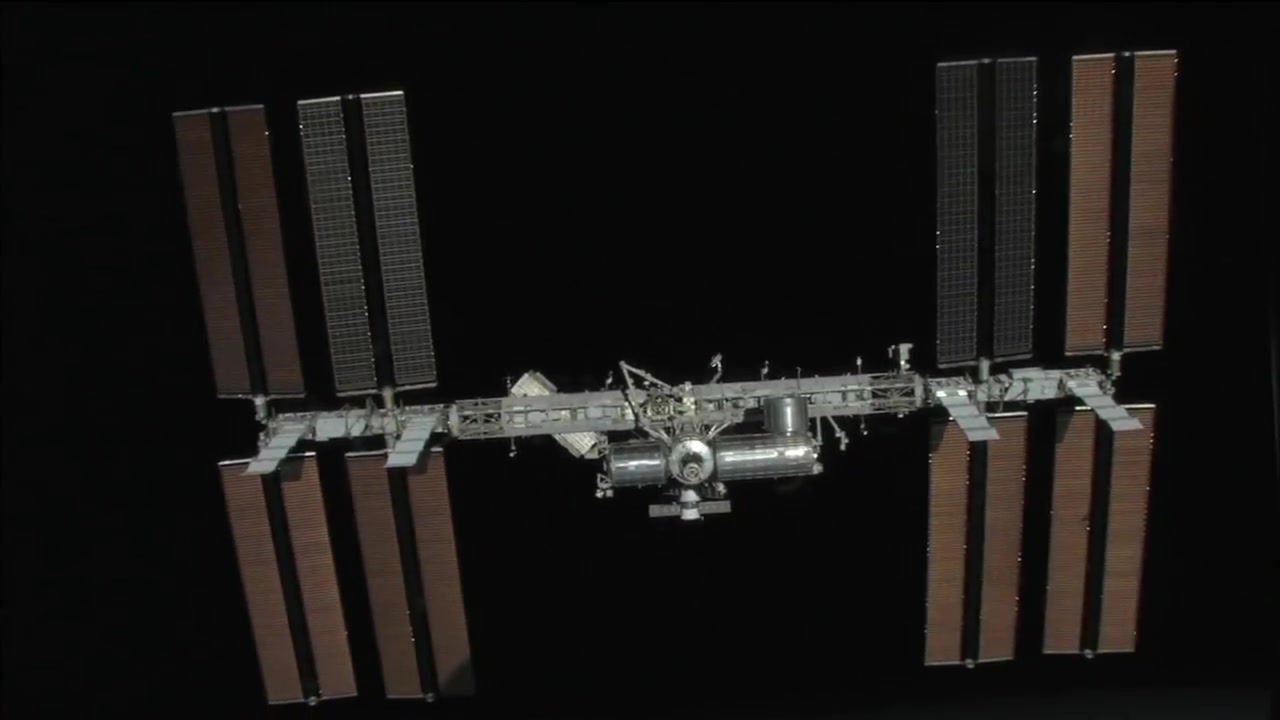}
    \end{subfigure}
    \hfill
    \begin{subfigure}[b]{0.32\textwidth}
        \includegraphics[width=\textwidth]{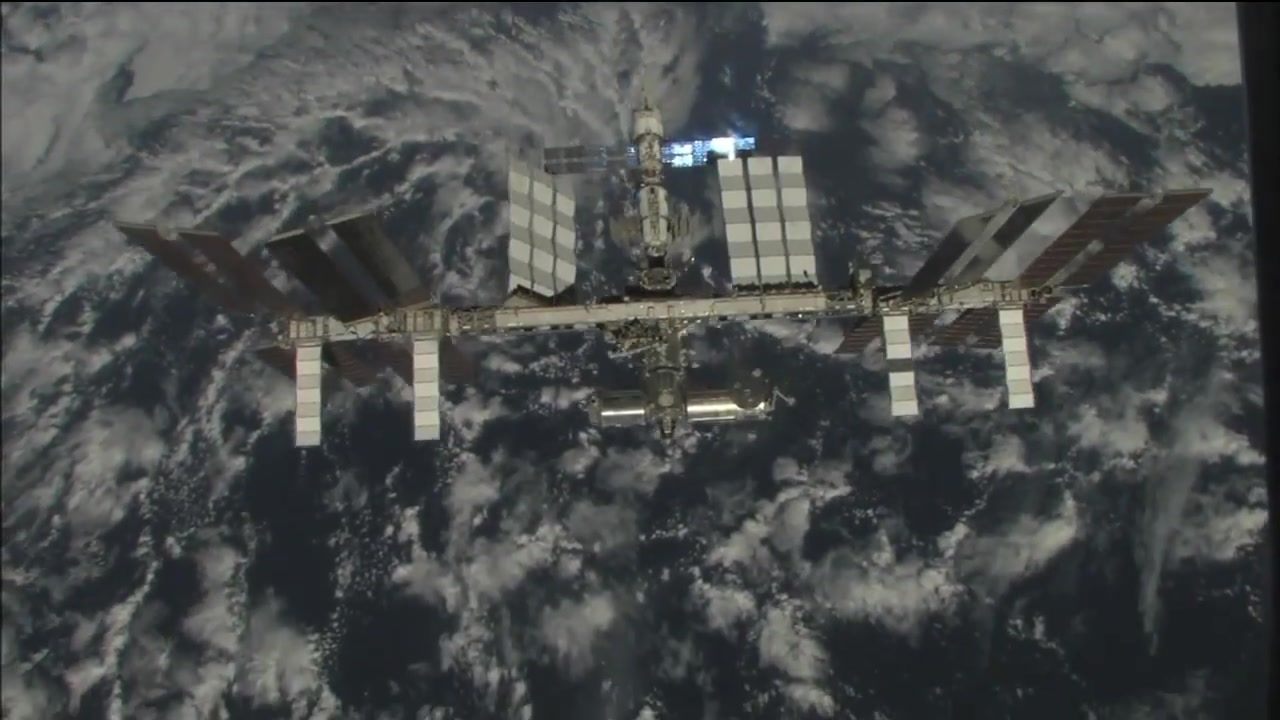}
    \end{subfigure}
    \hfill
    \begin{subfigure}[b]{0.32\textwidth}
        \includegraphics[width=\textwidth]{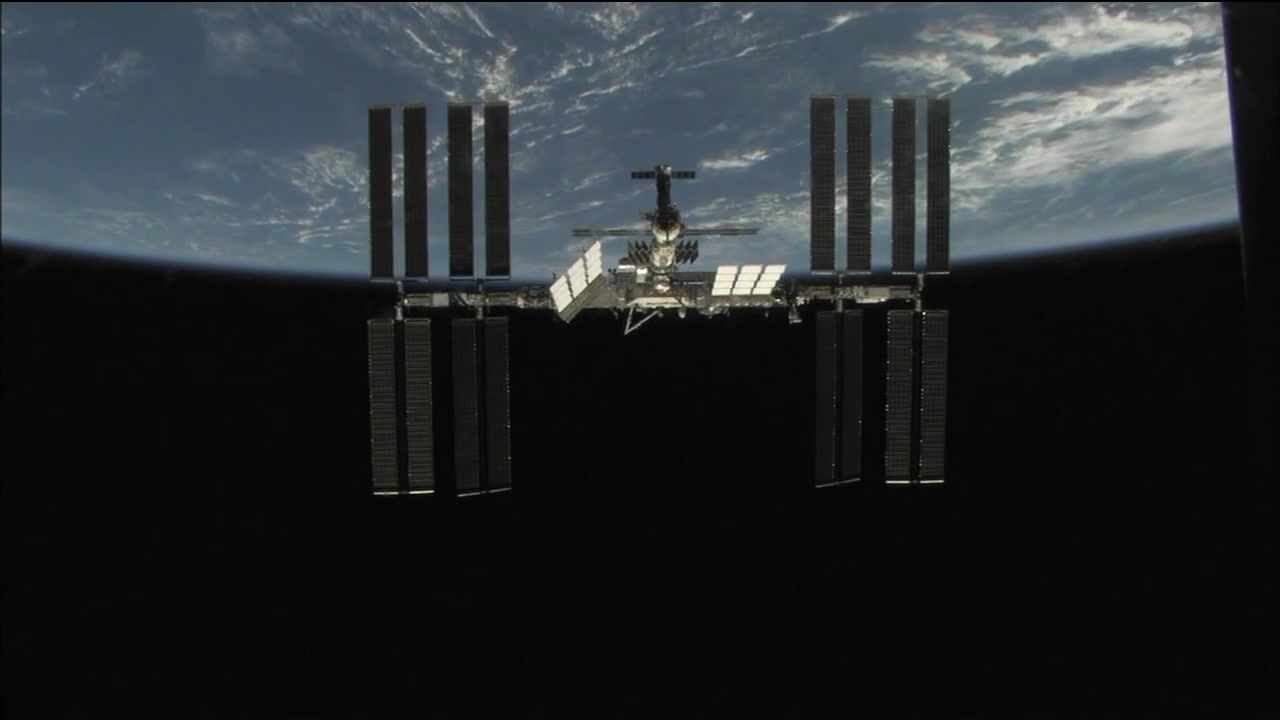}
    \end{subfigure}

    \vspace{0.5em}
    \begin{subfigure}[b]{0.32\textwidth}
        \includegraphics[width=\textwidth]{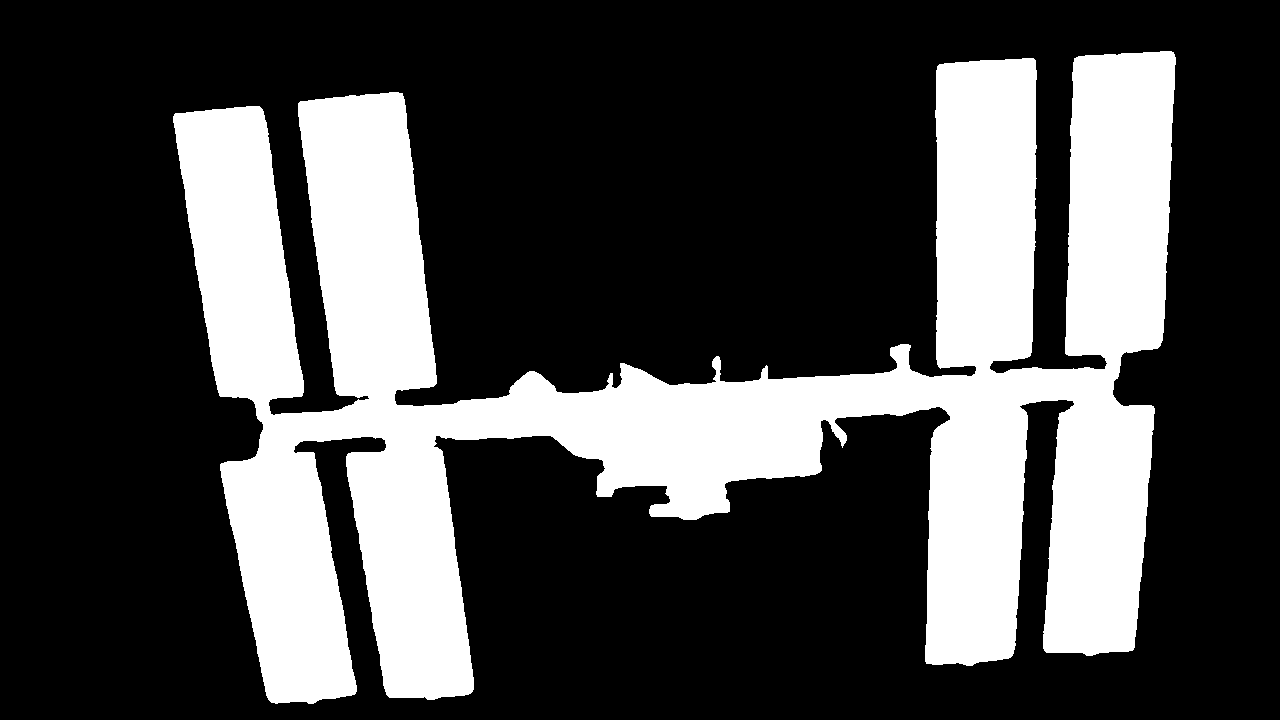}
    \end{subfigure}
    \hfill
    \begin{subfigure}[b]{0.32\textwidth}
        \includegraphics[width=\textwidth]{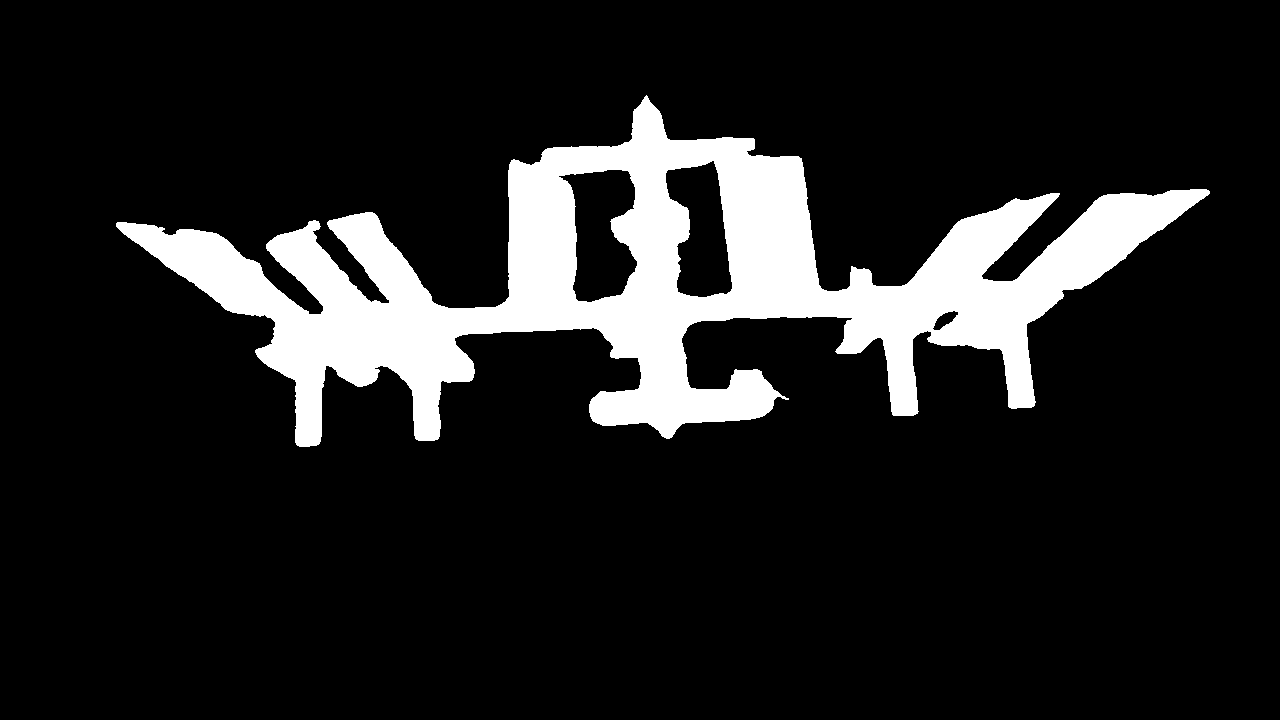}
    \end{subfigure}
    \hfill
    \begin{subfigure}[b]{0.32\textwidth}
        \includegraphics[width=\textwidth]{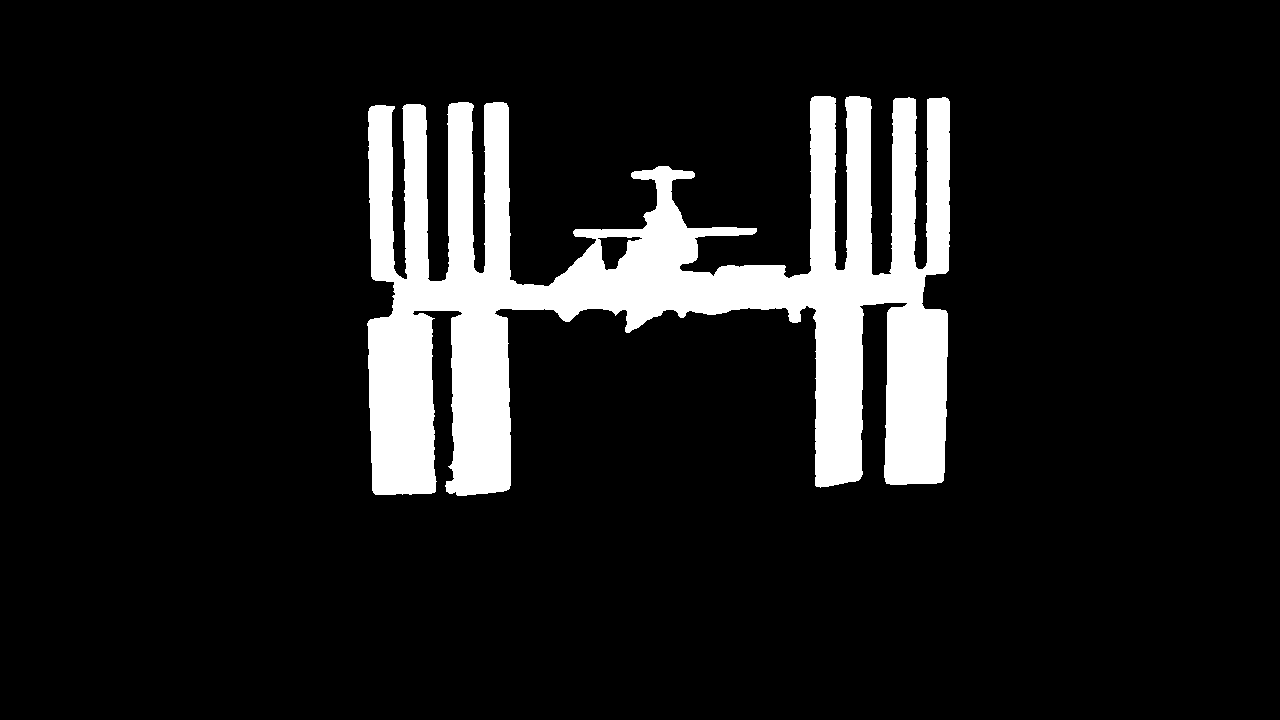}
    \end{subfigure}

    \vspace{0.5em}
    \begin{subfigure}[b]{0.32\textwidth}
        \includegraphics[width=\textwidth]{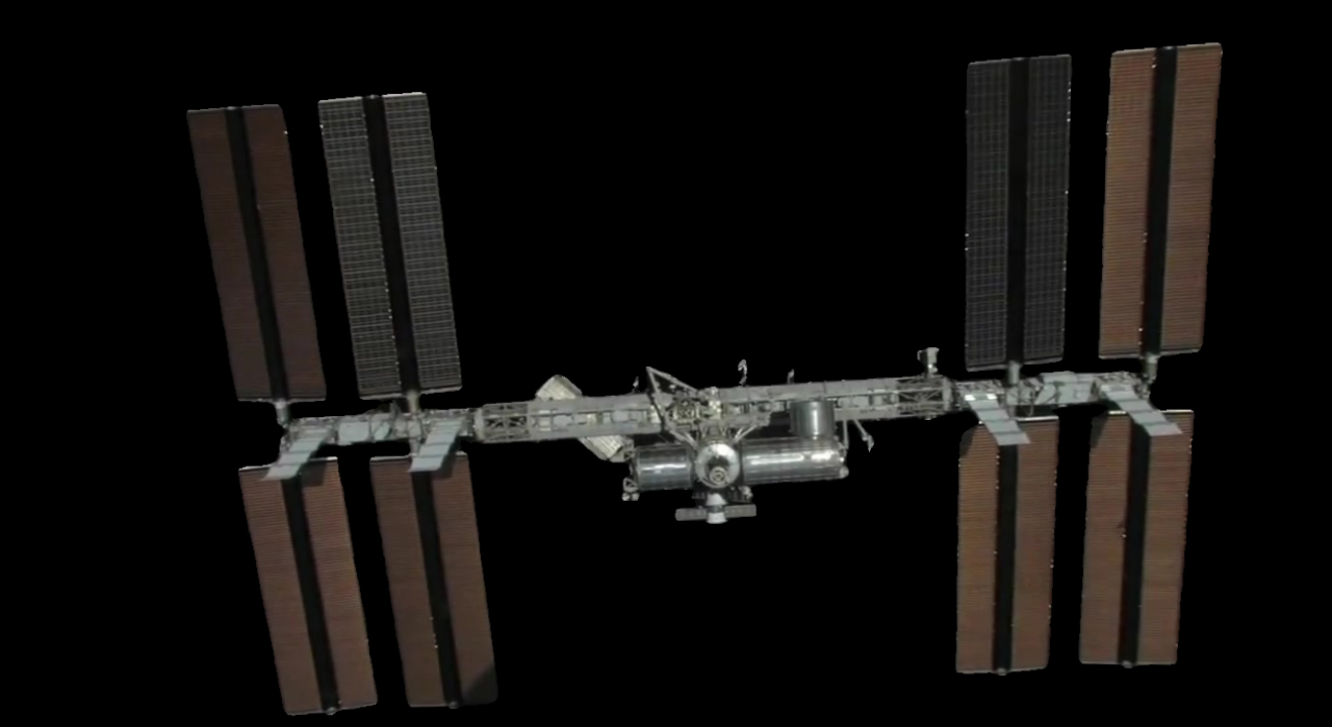}
    \end{subfigure}
    \hfill
    \begin{subfigure}[b]{0.32\textwidth}
        \includegraphics[width=\textwidth]{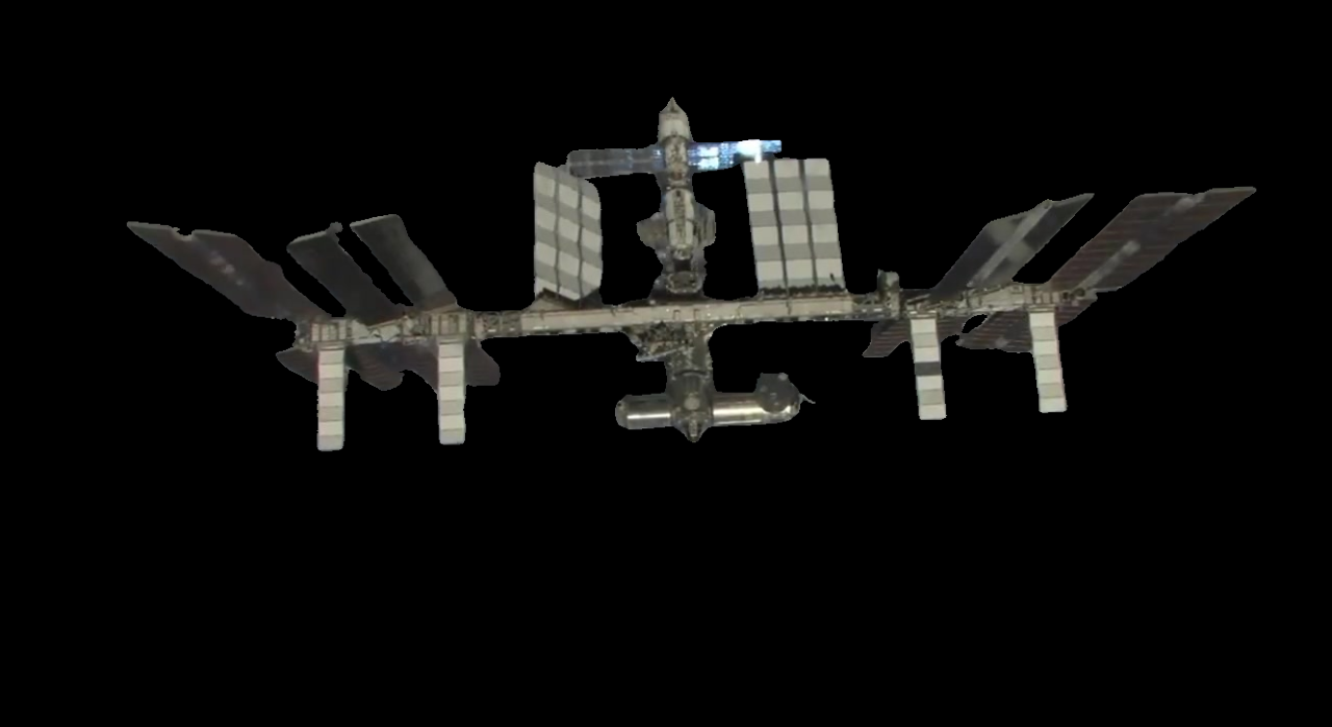}
    \end{subfigure}
    \hfill
    \begin{subfigure}[b]{0.32\textwidth}
        \includegraphics[width=\textwidth]{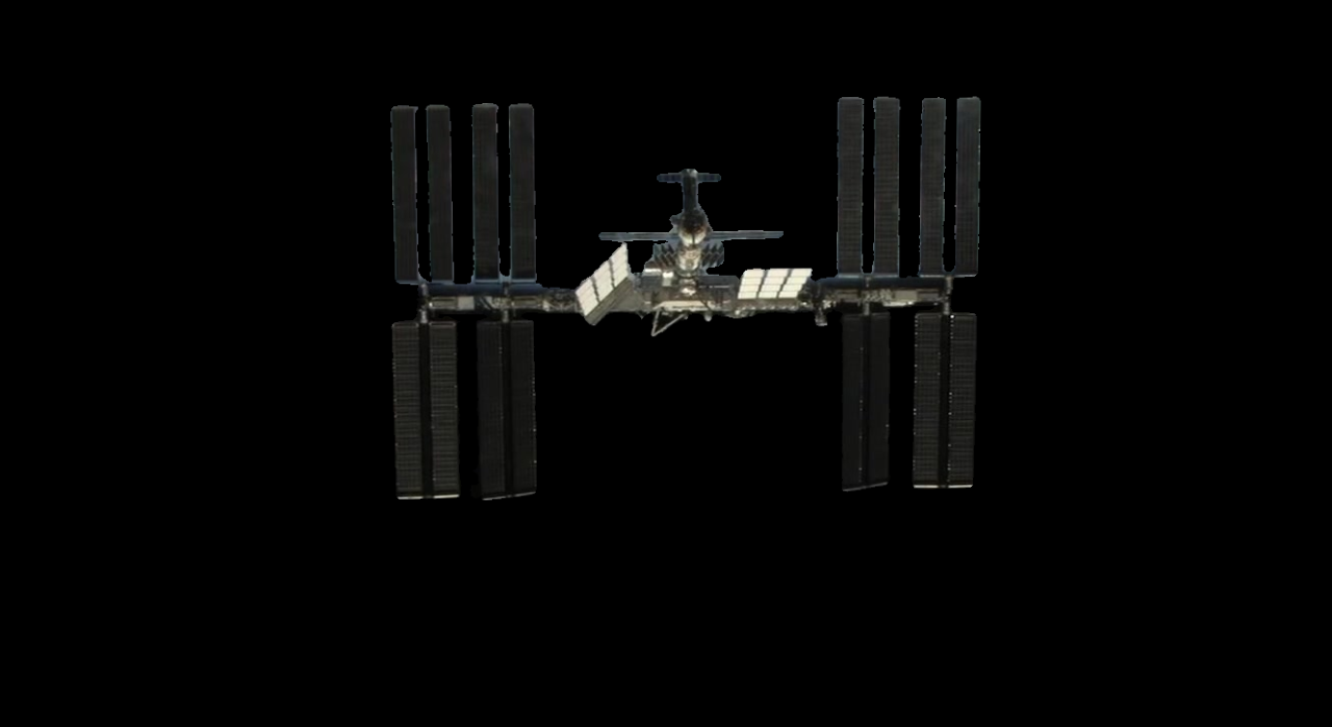}
    \end{subfigure}

    \caption{Sample frames from the publicly released STS-119 inspection video and their corresponding binary masks and background-removed frames generated by \samthree{}. Top row shows original frames, middle row shows the binary masks, and bottom row shows the segmented frames after background removal. The left column shows a frame with a black background, the center column shows a frame with Earth in the background, and the right column shows a frame during transition.}
    \label{fig:iss_images_masks_seg_images}
\end{figure}

We employed text-based prompting with \samthree{}, using "spacecraft" or "satellite" interchangeably as the prompt to extract binary masks for the upper stage rocket body, as \samthree{} requires the prompt to be a single-word noun. \samthree{} supports two propagation modes: image mode, in which each frame is segmented individually, and video mode, wherein the first frame is prompted and the mask is propagated across all frames automatically. We evaluated both image and video mode before selecting video mode for our pipeline based on empirical evaluation. Sample images and their corresponding masks and segmented frames are shown in Figure~\ref{fig:iss_images_masks_seg_images}. Once the masks are generated for each frame, pixels outside the mask are set to black.

\subsection{Structure from Motion with \colmap{}}
\label{sec:sfm}

With the background-removed frames, we run \colmap{} \cite{schoenberger2016sfm, schoenberger2016mvs} to recover the camera intrinsic matrix and per-frame poses. For video-based datasets captured during fly-around maneuvers, sequential matching is the natural choice. Based on findings from our initial experiments with both sequential and exhaustive matchers, we set the overlap parameter to 30 frames for feature matching. Although a larger overlap parameter increases processing time, we empirically found that an overlap of 30 frames yielded sufficient feature matches across images.

In contrast to the poor registration observed on the raw frames, background removal enabled successful registration across all video-based datasets, with dataset-specific registration results and camera intrinsics reported in Section~\ref{sec:sfm_results}.


The recovered camera poses and intrinsics are subsequently passed to \neuralangelo{} as input for surface reconstruction.

\subsection{Neural Implicit Surface Reconstruction via \neuralangelo{}}
\label{sec:neuralangelo}

\neuralangelo{} \cite{li2023neuralangelo} is a neural implicit surface reconstruction method that represents the scene as a signed distance function (SDF) and optimizes it using multi-resolution hash encodings. Given the camera poses and intrinsics estimated by \colmap{}, we run \neuralangelo{} on the background-removed frames to recover the 3D surface of the object of interest. The output of \neuralangelo{} is a learned SDF, from which a mesh is extracted using the Marching Cubes algorithm at a specified resolution. Compared to standard neural implicit methods such as NeRF, \neuralangelo{} is better suited for recovering fine surface details because it progressively refines the surface from coarse to fine resolution levels during training, capturing details that standard methods fail to resolve. However, on-orbit inspection imagery exhibits significant photometric variations due to shadows and changing illumination conditions. Since \neuralangelo{} assumes photometric consistency across views, these variations can adversely affect reconstruction fidelity, as they are baked into the texture of the reconstructed mesh. This limitation motivates the use of a post-processing method designed specifically to handle such photometric variations.

\subsection{Photometric Post-Processing via \ppisp{}}
\label{sec:ppisp}

As noted in Section~\ref{sec:neuralangelo}, on-orbit inspection imagery exhibits significant photometric variation due to shadows and changing illumination conditions, which can become baked into the texture of the reconstructed mesh. To address this, we incorporate \ppisp{}, a physically-grounded correction module that corrects photometric inconsistencies in multi-view 3D reconstruction by disentangling camera-intrinsic effects from capture-dependent variations. During the first training phase, \ppisp{} optimizes a sequence of physically-based modules applied to the reconstructed radiance, namely exposure offset, chromatic vignetting, linear color correction, and a non-linear camera response function, alongside \neuralangelo{} for 400k iterations. In the second phase, the \neuralangelo{} parameters are frozen and the \ppisp{} controller is trained for the remaining 100k iterations to predict per-frame exposure and color correction parameters. Qualitative comparisons between meshes generated with and without \ppisp{} across datasets are presented in Section~\ref{sec:PPISP_evaluation}.

\section{Experimental Evaluation}
\label{sec:experiments}
In this section, we present our experimental setup, dataset details, and the results obtained from our pipeline.

\subsection{Dataset}
\label{sec:dataset}




\subsubsection{STS-119}
\label{sec:dataset_sts119}

The STS-119 dataset consists of publicly released footage from NASA's STS-119 mission, in which the Space Shuttle Discovery performed a fly-around of the International Space Station (ISS) in March 2009 \cite{nasa2009sts119}. The footage was captured at 59.94 frames per second (FPS) with a total duration of approximately 168.73 seconds, yielding 10,114 frames at a resolution of $1280 \times 720$ pixels. We extract 506 frames using a downsampling factor of 20, and rely on qualitative evaluation of the reconstructed mesh.

\subsubsection{ADRAS-J}
\label{sec:dataset_adrasj}

The ADRAS-J dataset consists of publicly released on-orbit inspection video captured by \adrasj{}, a space debris inspection spacecraft developed by Astroscale Japan and selected by JAXA for Phase I of its Commercial Removal of Debris Demonstration Project (CRD2). The mission objective was to safely perform rendezvous and proximity operations (RPO) and to characterize and assess the condition of an H-IIA upper stage rocket body left in low Earth orbit in 2009. As part of the mission, \adrasj{} performed a fly-around maneuver around the upper stage on July 15 and 16, 2024, at a standoff distance of approximately 50 meters, capturing imagery from various angles and illumination conditions \cite{jaxa2024crd2, astroscale2024flyaround}.

The inspection video used in this work was captured at 24 frames per second (FPS) with a total duration of approximately 34.92 seconds, yielding 838 frames at a resolution of $720 \times 960$ pixels (portrait orientation, 3:4 aspect ratio) \cite{astroscale2024video}. For our experiments, we extract 419 frames using a downsampling factor of 2, consistent with the image count used in our prior work \cite{gopu2026dynamic}.

As no official ground truth CAD model of the H-IIA upper stage is publicly available, only artist-created approximations exist, and we therefore rely on qualitative evaluation of the reconstructed mesh.

\subsection{Implementation Details}

We implemented our pipeline on a workstation equipped with two NVIDIA RTX 4090 GPUs, each with 24 GB of VRAM. To accelerate the segmentation process, we leveraged the multi-GPU support of \samthree{} by distributing the model across both GPUs. Model loading was consistent across datasets, \samthree{} utilized 4.02 GB and 3.98 GB of peak VRAM on GPU 0 and GPU 1 respectively, taking approximately 19-22 seconds. Inference times and VRAM usage varied with the number of frames processed. For the STS-119 dataset (506 frames), at inference, \samthree{} utilized 7.17 GB on GPU 0 and 6.12 GB on GPU 1, taking 78.24 seconds. For the \adrasj{} dataset (419 frames), at inference, \samthree{} utilized 6.21 GB on GPU 0 and 5.40 GB on GPU 1, taking 56.50 seconds. For \colmap{}, we used the scripts and \colmap{} binary bundled within the \neuralangelo{} source code. \neuralangelo{} training is based on a fixed number of iterations rather than epochs or number of training images; we used the default of 500,000 iterations. Training and mesh generation took approximately 9 hours on a single GPU. 

\subsection{Structure from Motion Results}
\label{sec:sfm_results}

\begin{figure}[H]
    \centering
    \begin{subfigure}[b]{0.48\linewidth}
        \includegraphics[width=\textwidth]{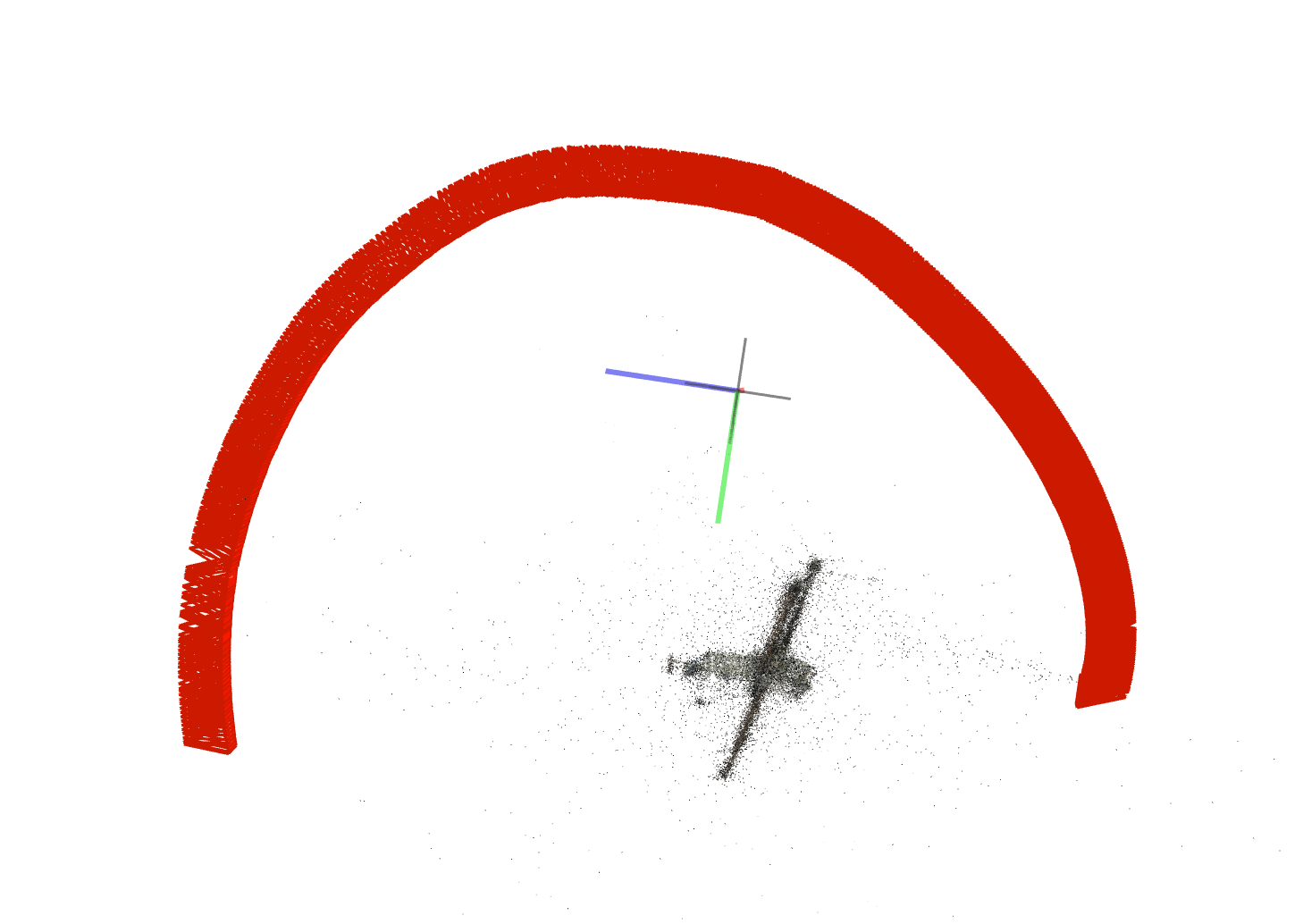}
        \caption{STS-119}
        \label{fig:colmap_sts119}
    \end{subfigure}
    \hfill
    \begin{subfigure}[b]{0.48\linewidth}
        \includegraphics[width=\textwidth]{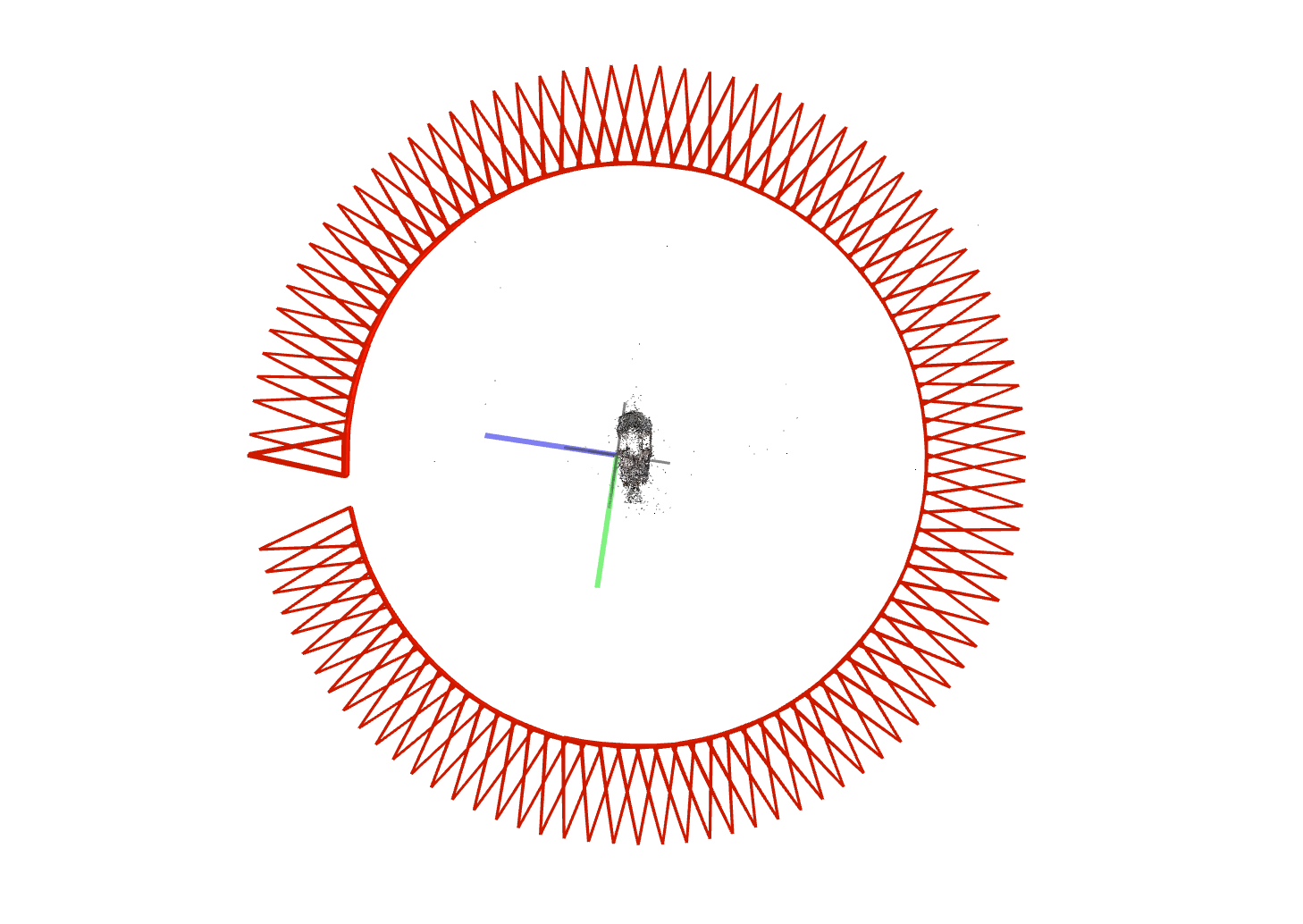}
        \caption{ADRAS-J}
        \label{fig:colmap_adrasj}
    \end{subfigure}
    \caption{Recovered camera trajectories (red frustums) and sparse point 
    clouds from \colmap{} for the STS-119 and ADRAS-J datasets.}
    \label{fig:colmap_results}
\end{figure}

After \samthree{} background removal, \colmap{} successfully registered all frames for both the STS-119 and \adrasj{} datasets. The registration results are summarized in Table~\ref{tab:colmap_results}. As shown in Figure~\ref{fig:colmap_results}, the recovered camera trajectory for STS-119 forms a semicircle around the ISS, reflecting the partial nature of the fly-around, while the \adrasj{} trajectory forms a near-complete circle around the H-IIA upper stage, consistent with the full fly-around maneuver. To further validate the recovered \adrasj{} camera poses, we imported the reconstructed mesh and camera poses into Blender and overlaid the segmented inspection frames onto the corresponding camera views. The visual alignment between the mesh geometry and the segmented imagery, achieved without any manual rotation or alignment, confirms the accuracy of the recovered camera trajectory.

\begin{table}[h]
    \centering
    \begin{tabular}{lcc}
        \toprule
        & STS-119 & ADRAS-J \\
        \midrule
        Registered frames & 506 / 506 & 419 / 419 \\
        Sparse point cloud & 35,740 & 16,505 \\
        Mean reprojection error (px) & 0.664 & 0.410 \\
        Focal length (px) & 1147.89 & 2396.08 \\
        Principal point (px) & (669.0, 364.0) & (364.5, 492.5) \\
        Estimated dimensions (px) & $1338 \times 728$ & $729 \times 985$ \\
        Nominal dimensions (px) & $1280 \times 720$ & $720 \times 960$ \\
        \bottomrule
    \end{tabular}
    \caption{\colmap{} registration results for the STS-119 and ADRAS-J datasets.}
    \label{tab:colmap_results}
\end{table}

\subsection{Reconstruction Results}
\label{sec:reconstruction_results}

\begin{figure}[H]
    \centering
    \includegraphics[width=0.75\linewidth]{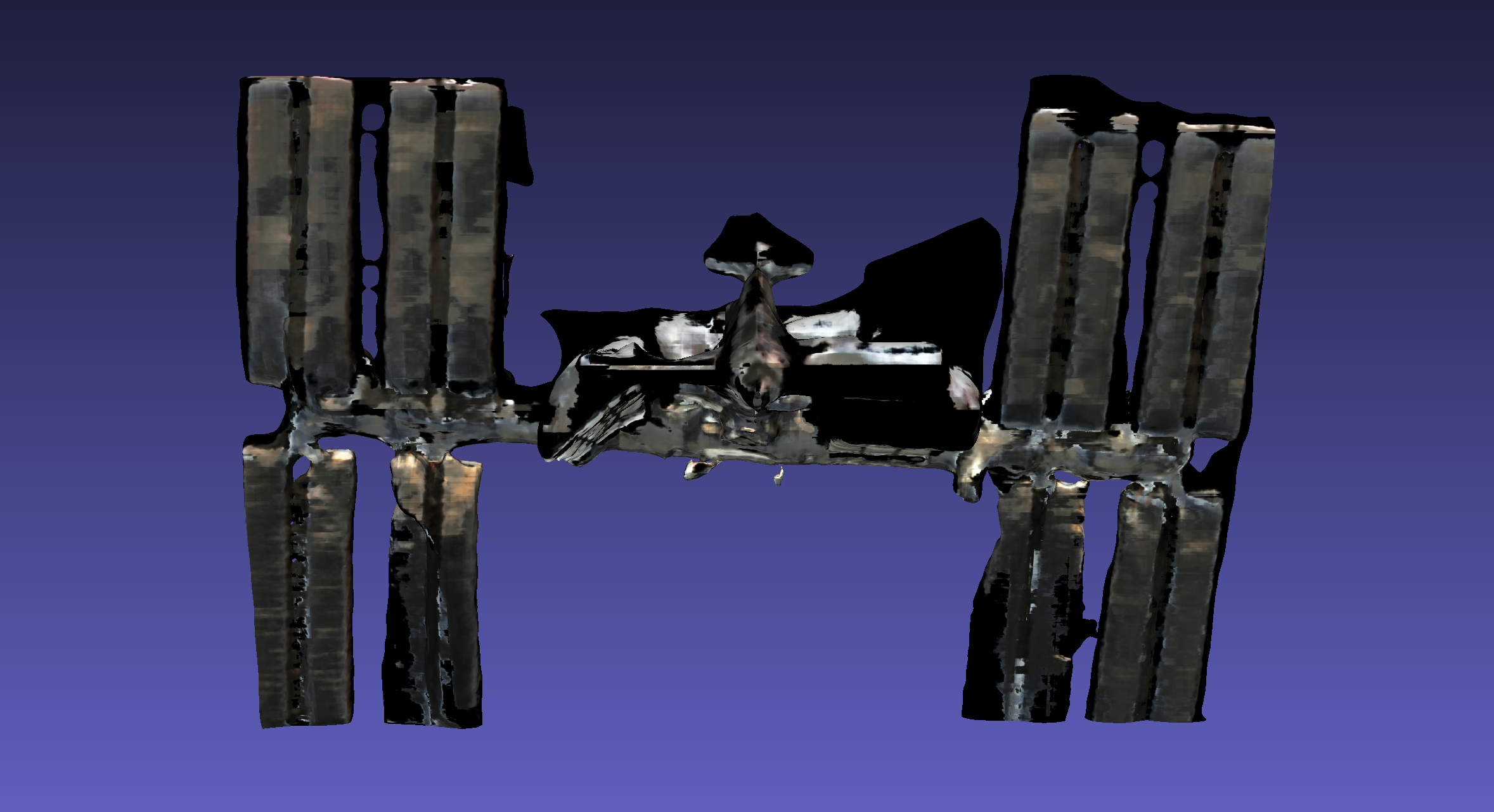}
    \caption{Reconstructed mesh of the ISS generated by \neuralangelo{} without \ppisp{} post-processing, viewed in MeshLab \cite{cignoni2008meshlab}.}
    \label{fig:iss_mesh}
\end{figure}

\begin{figure}[H]
    \centering
    \includegraphics[width=0.5\linewidth]{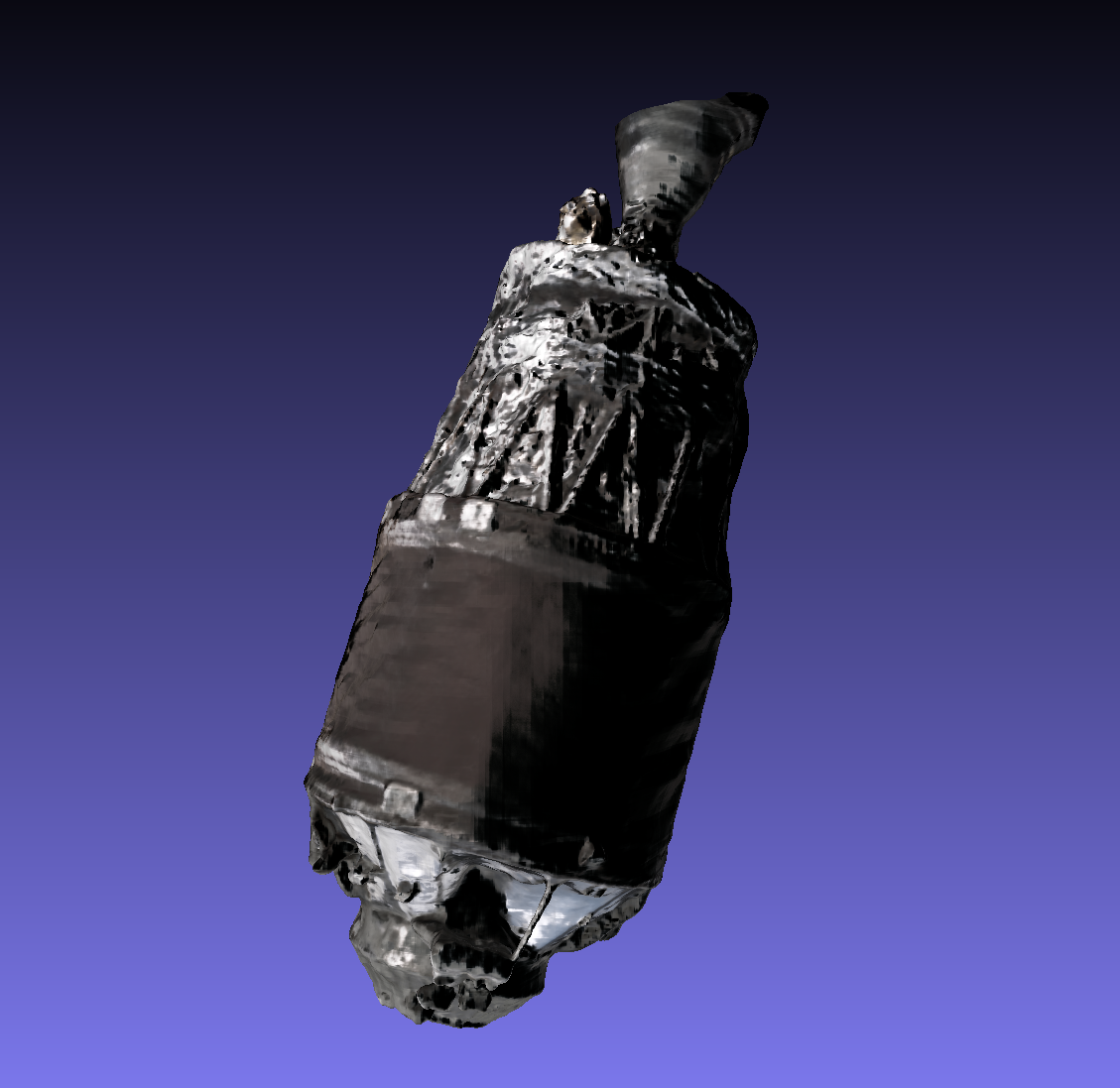}
    \caption{Reconstructed mesh of the H-IIA upper stage generated by \neuralangelo{} without \ppisp{} post-processing, viewed in MeshLab \cite{cignoni2008meshlab}.}
    \label{fig:upperstage_mesh}
\end{figure}

The reconstructed meshes of the ISS and the H-IIA upper stage generated by \neuralangelo{} are shown in Figures~\ref{fig:iss_mesh} and \ref{fig:upperstage_mesh}, respectively, both extracted using the Marching Cubes algorithm at a resolution of 2048, yielding 5,282,218 vertices and 10,395,612 faces for the ISS and 5,953,279 vertices and 11,717,890 faces for the H-IIA upper stage.

\begin{figure}[H]
    \centering
    \includegraphics[width=\linewidth]{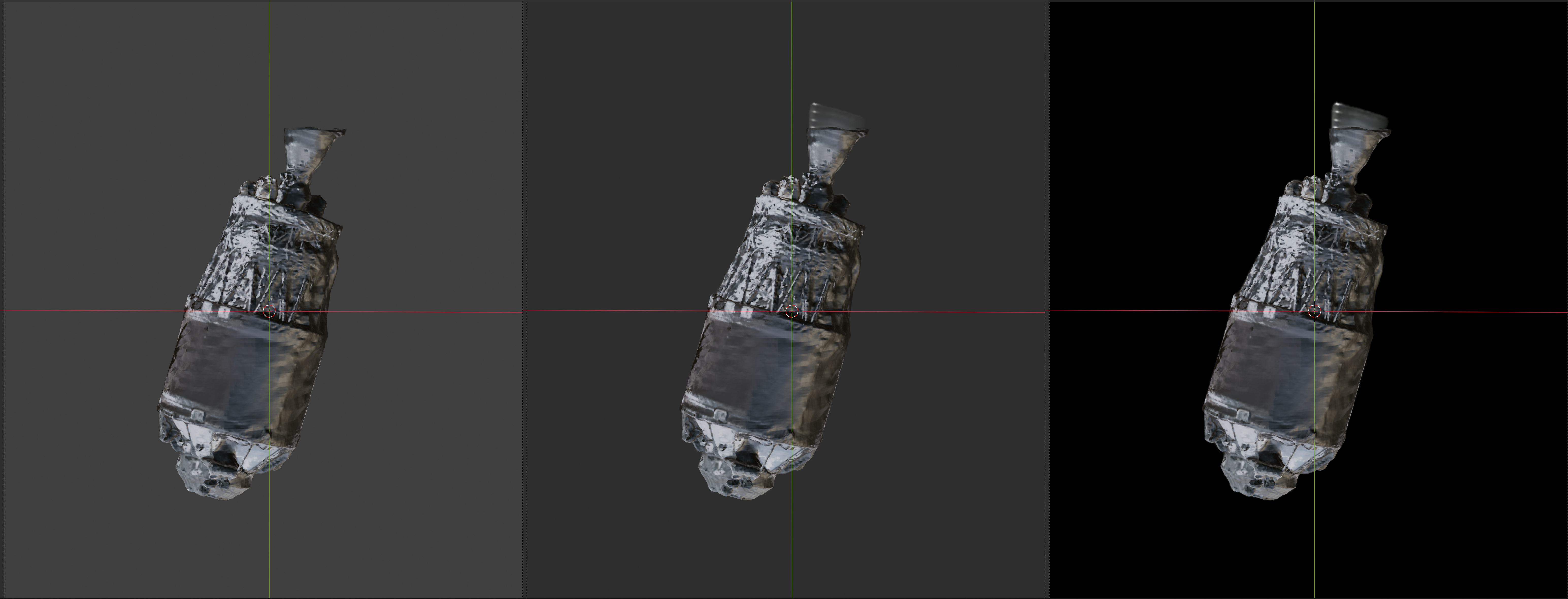}
    \caption{Reprojection of the reconstructed mesh of the H-IIA upper stage without \ppisp{} overlaid with the segmented inspection image at opacity 0.0 (left), 0.5 (center), and 1.0 (right), viewpoint 1.}
    \label{fig:reprojection_view1}
\end{figure}


\subsection{PPISP Evaluation}
\label{sec:PPISP_evaluation}

\begin{figure}
    \centering
    \includegraphics[width=\linewidth]{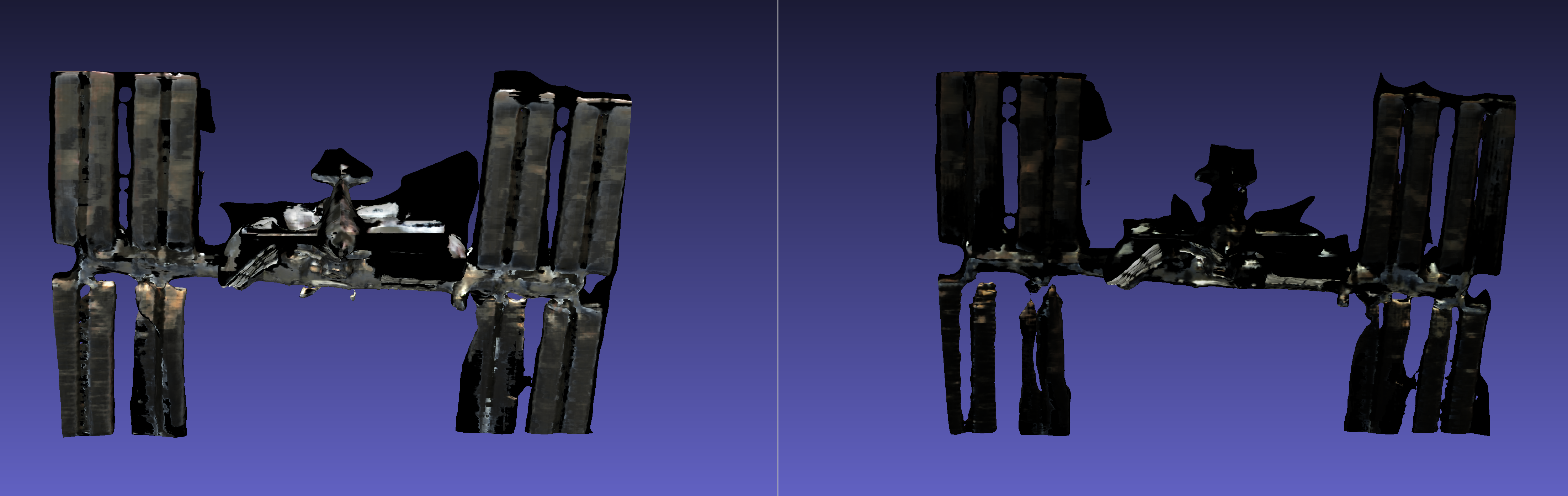}
    \caption{Qualitative comparison of the reconstructed mesh of the ISS without \ppisp{} (left) and with \ppisp{} (right), viewpoint 1.}
    \label{fig:iss_ppisp_comparison_viewpoint1}
\end{figure}
\begin{figure}
    \centering
    \includegraphics[width=\linewidth]{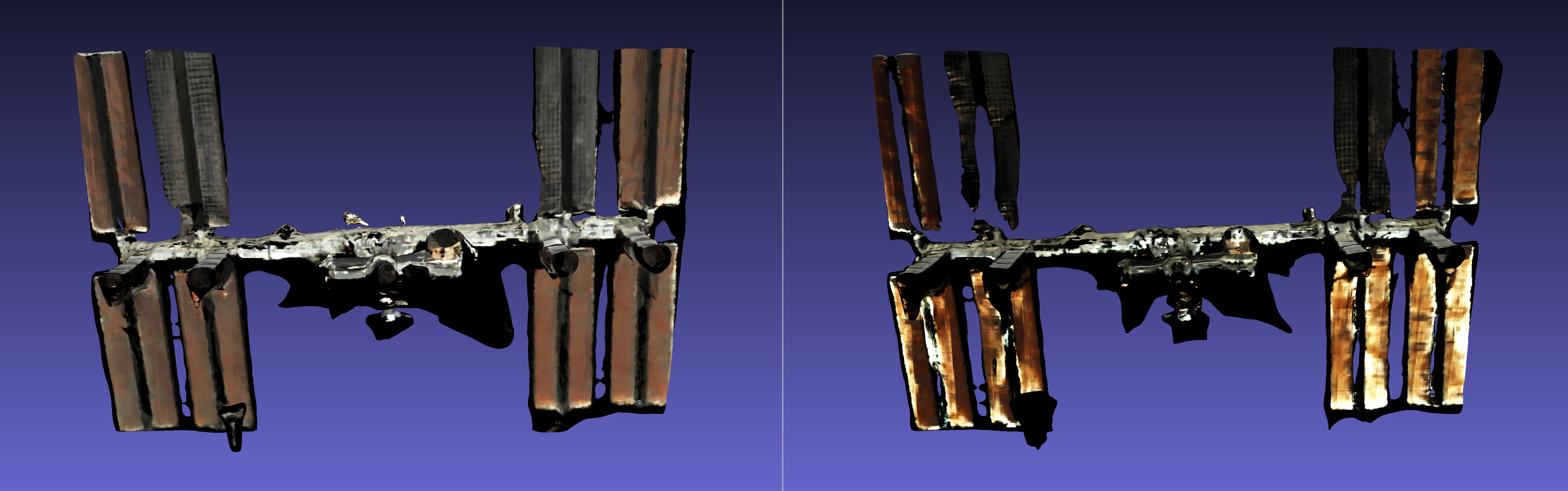}
    \caption{Qualitative comparison of the reconstructed mesh of the ISS without \ppisp{} (left) and with \ppisp{} (right), viewpoint 2.}
    \label{fig:iss_ppisp_comparison_viewpoint2}
\end{figure}
\begin{figure}
    \centering
    \includegraphics[width=\linewidth]{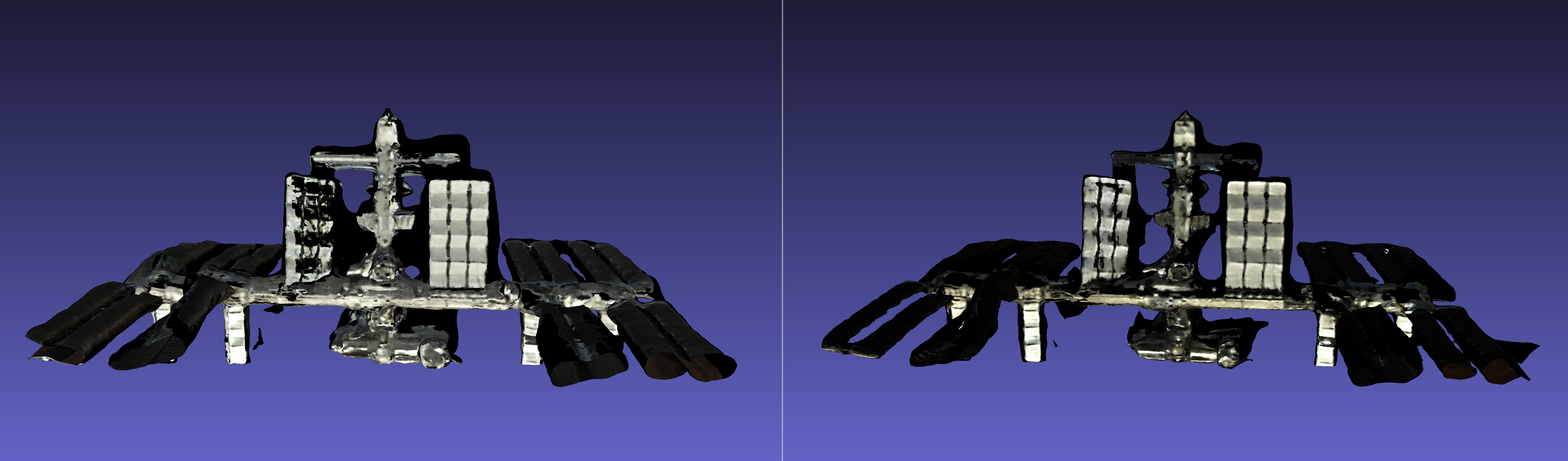}
    \caption{Qualitative comparison of the reconstructed mesh of the ISS without \ppisp{} (left) and with \ppisp{} (right), viewpoint 3.}
    \label{fig:iss_ppisp_comparison_viewpoint3}
\end{figure}
\begin{figure}
    \centering
    \includegraphics[width=\linewidth]{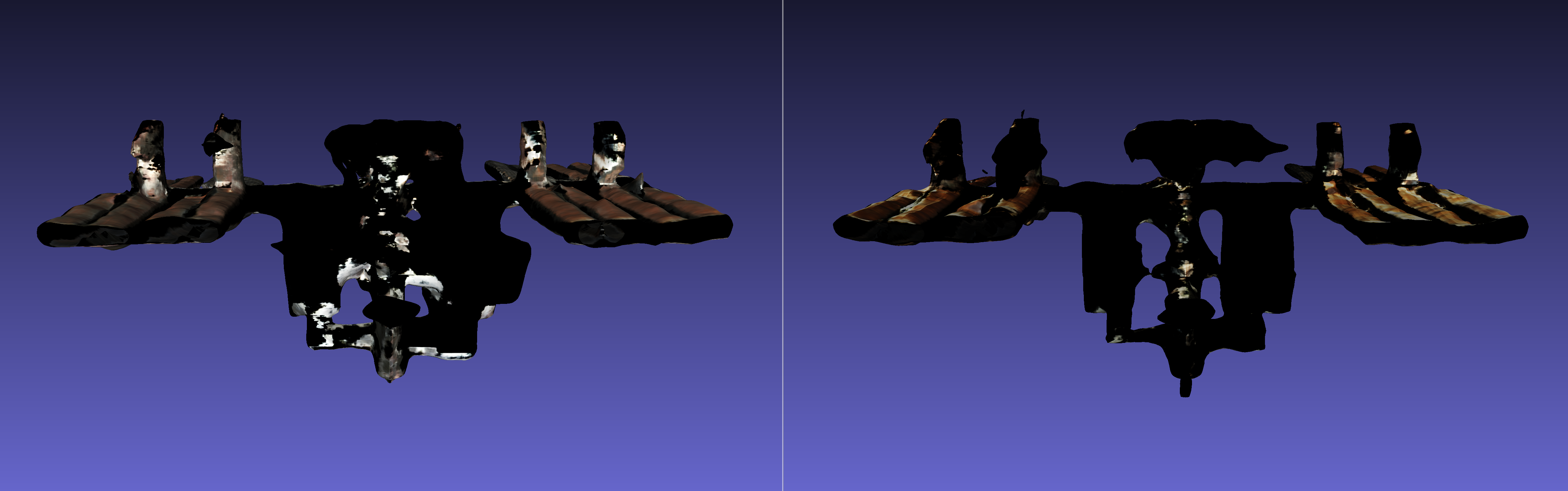}
    \caption{Qualitative comparison of the reconstructed mesh of the ISS without \ppisp{} (left) and with \ppisp{} (right), viewpoint 4.}
    \label{fig:iss_ppisp_comparison_viewpoint4}
\end{figure}
\begin{figure}
    \centering
    \includegraphics[width=\linewidth]{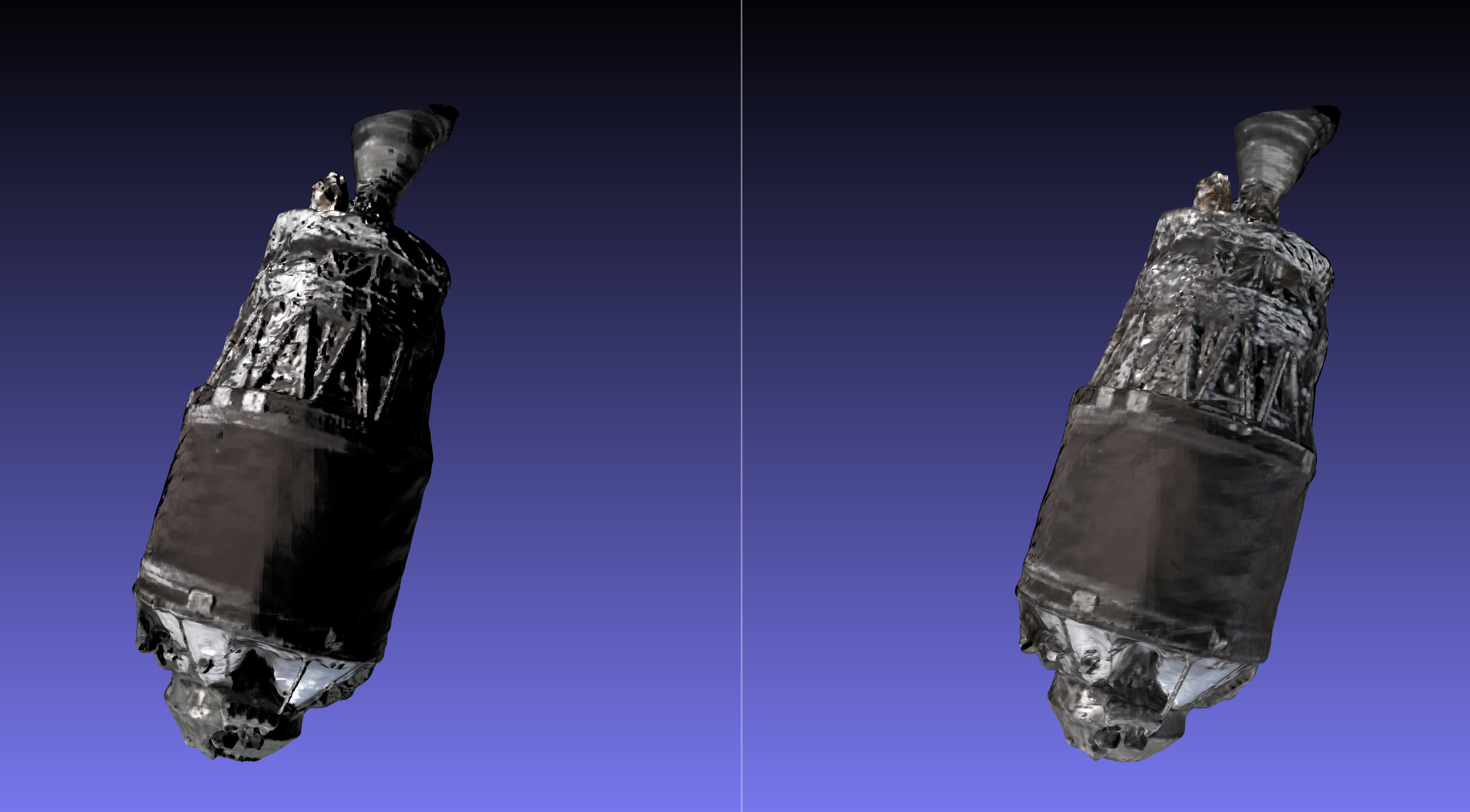}
    \caption{Qualitative comparison of the reconstructed mesh of the H-IIA upper stage without \ppisp{} (left) and with \ppisp{} (right), viewpoint 1.}
    \label{fig:ppisp_comparison_viewpoint_1}
\end{figure}
\begin{figure}
    \centering
    \includegraphics[width=\linewidth]{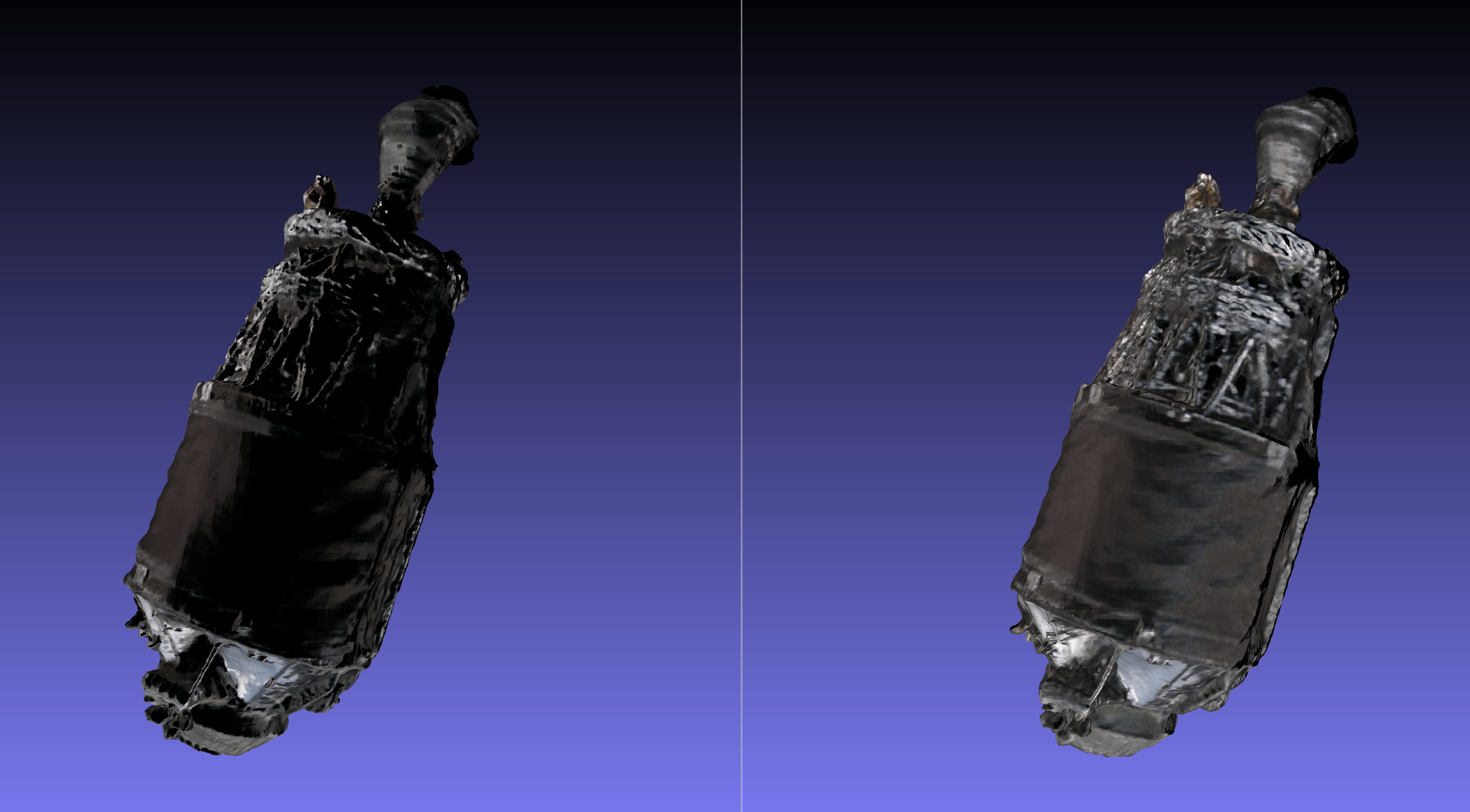}
    \caption{Qualitative comparison of the reconstructed mesh of the H-IIA upper stage without \ppisp{} (left) and with \ppisp{} (right), viewpoint 2.}
    \label{fig:ppisp_comparison_viewpoint_2}
\end{figure}
\begin{figure}
    \centering
    \includegraphics[width=\linewidth]{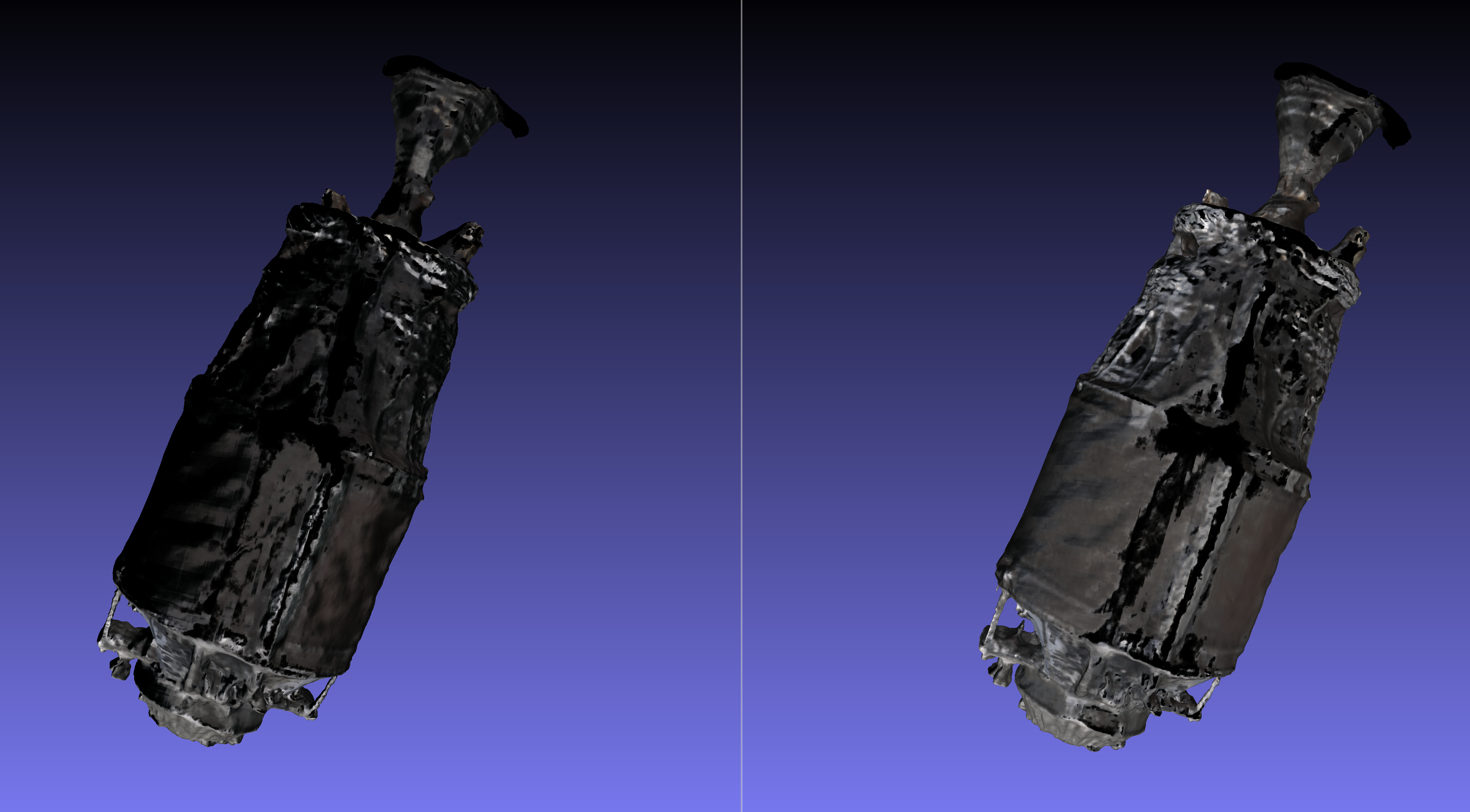}
    \caption{Qualitative comparison of the reconstructed mesh of the H-IIA upper stage without \ppisp{} (left) and with \ppisp{} (right), viewpoint 3.}
    \label{fig:ppisp_comparison_viewpoint_3}
\end{figure}
\begin{figure}
    \centering
    \includegraphics[width=\linewidth]{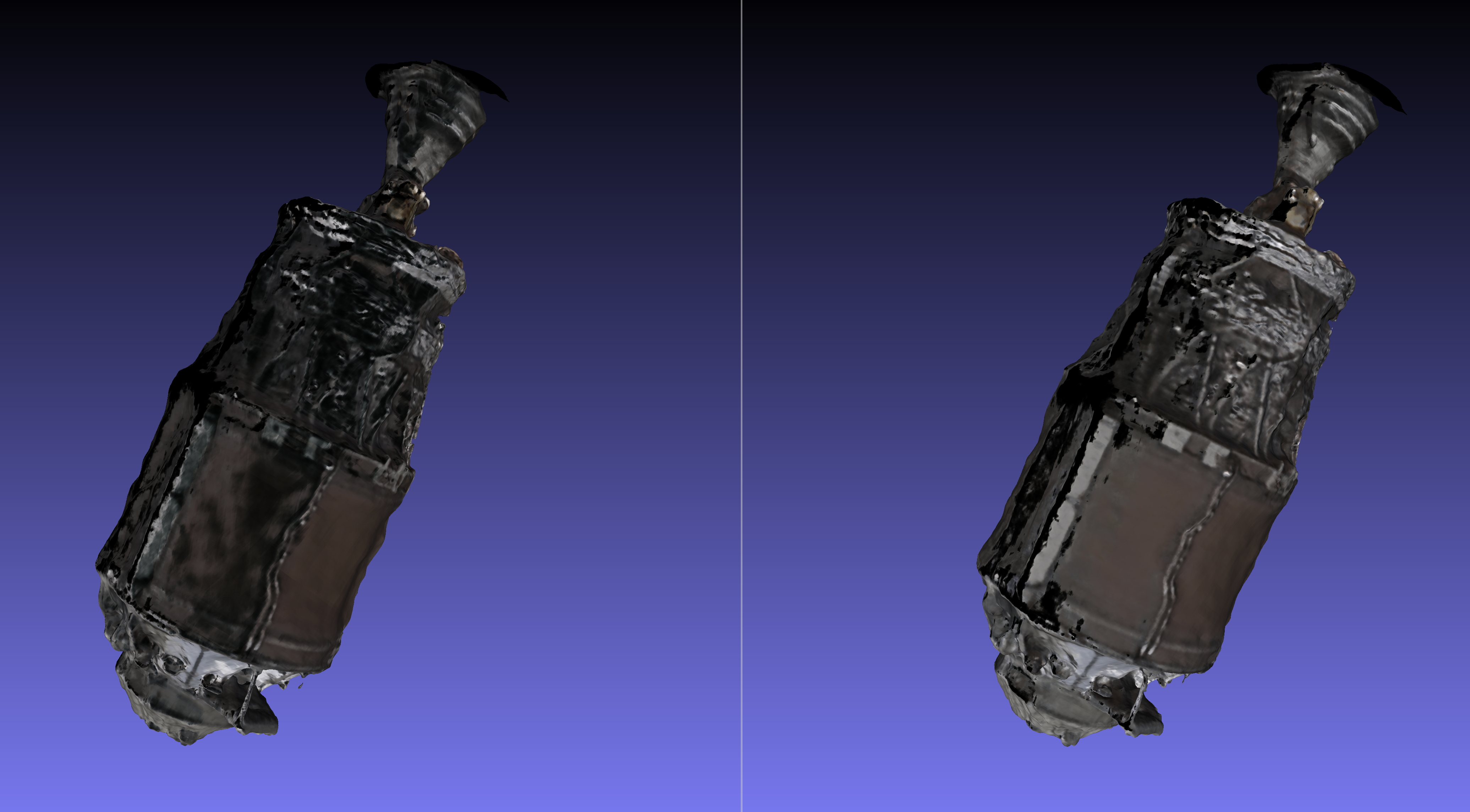}
    \caption{Qualitative comparison of the reconstructed mesh of the H-IIA upper stage without \ppisp{} (left) and with \ppisp{} (right), viewpoint 4.}
    \label{fig:ppisp_comparison_viewpoint_4}
\end{figure}

On-orbit inspection imagery inherently exhibits photometric variations across frames due to changing illumination conditions, which become embedded in the texture of the reconstructed mesh. To address these variations, we integrated \ppisp{} into the pipeline as described in Section~\ref{sec:ppisp}. A qualitative comparison of the reconstructed meshes with and without \ppisp{} for the ISS and the H-IIA upper stage is presented in Figures~\ref{fig:iss_ppisp_comparison_viewpoint1}--\ref{fig:iss_ppisp_comparison_viewpoint4} and \ref{fig:ppisp_comparison_viewpoint_1}--\ref{fig:ppisp_comparison_viewpoint_4}, respectively.

\begin{figure}
    \centering
    \includegraphics[width=\linewidth]{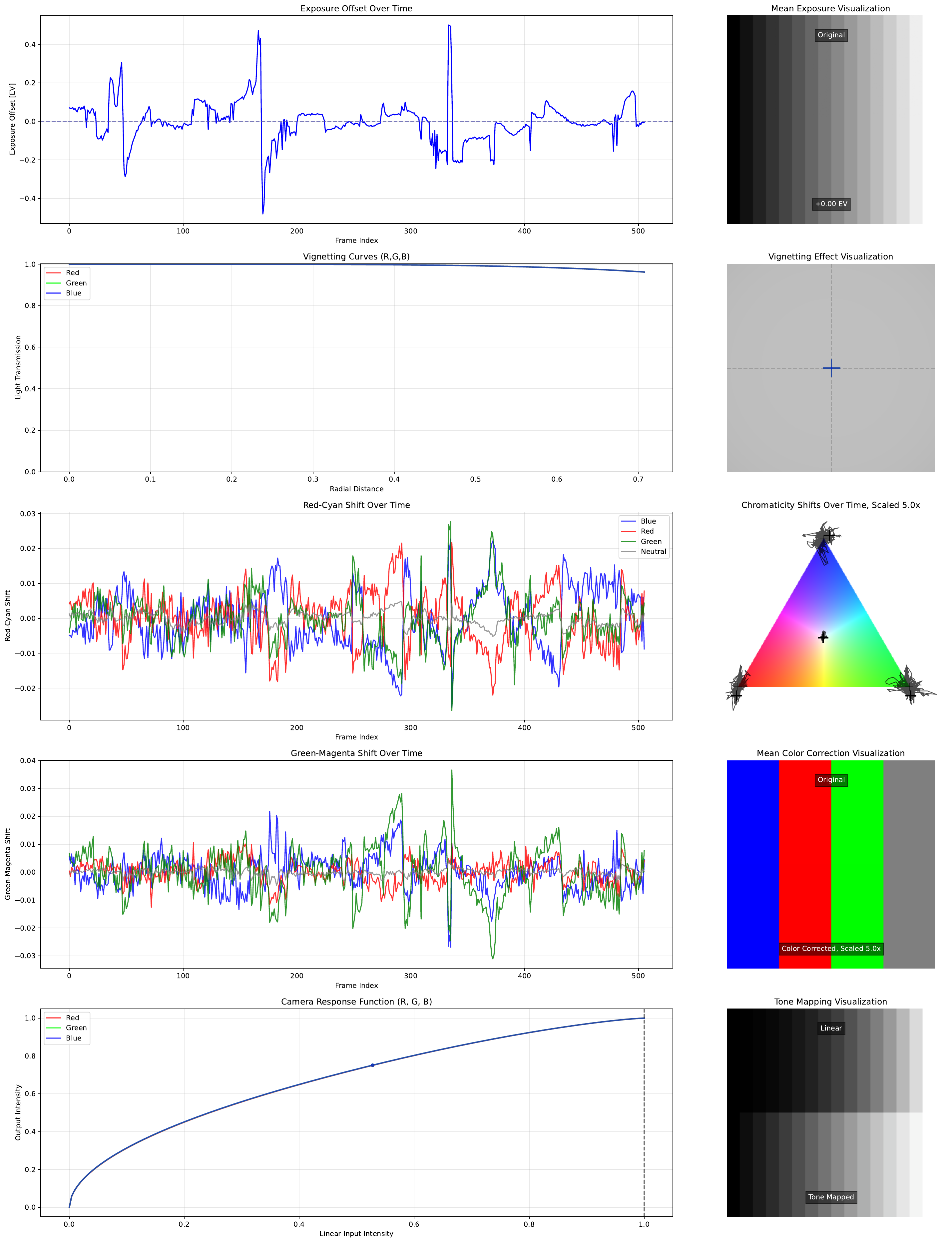}
    \caption{\ppisp{} evaluation report for the STS-119 dataset.}
    \label{fig:ppisp_report_iss}
\end{figure}

\begin{figure}
    \centering
    \includegraphics[width=\linewidth]{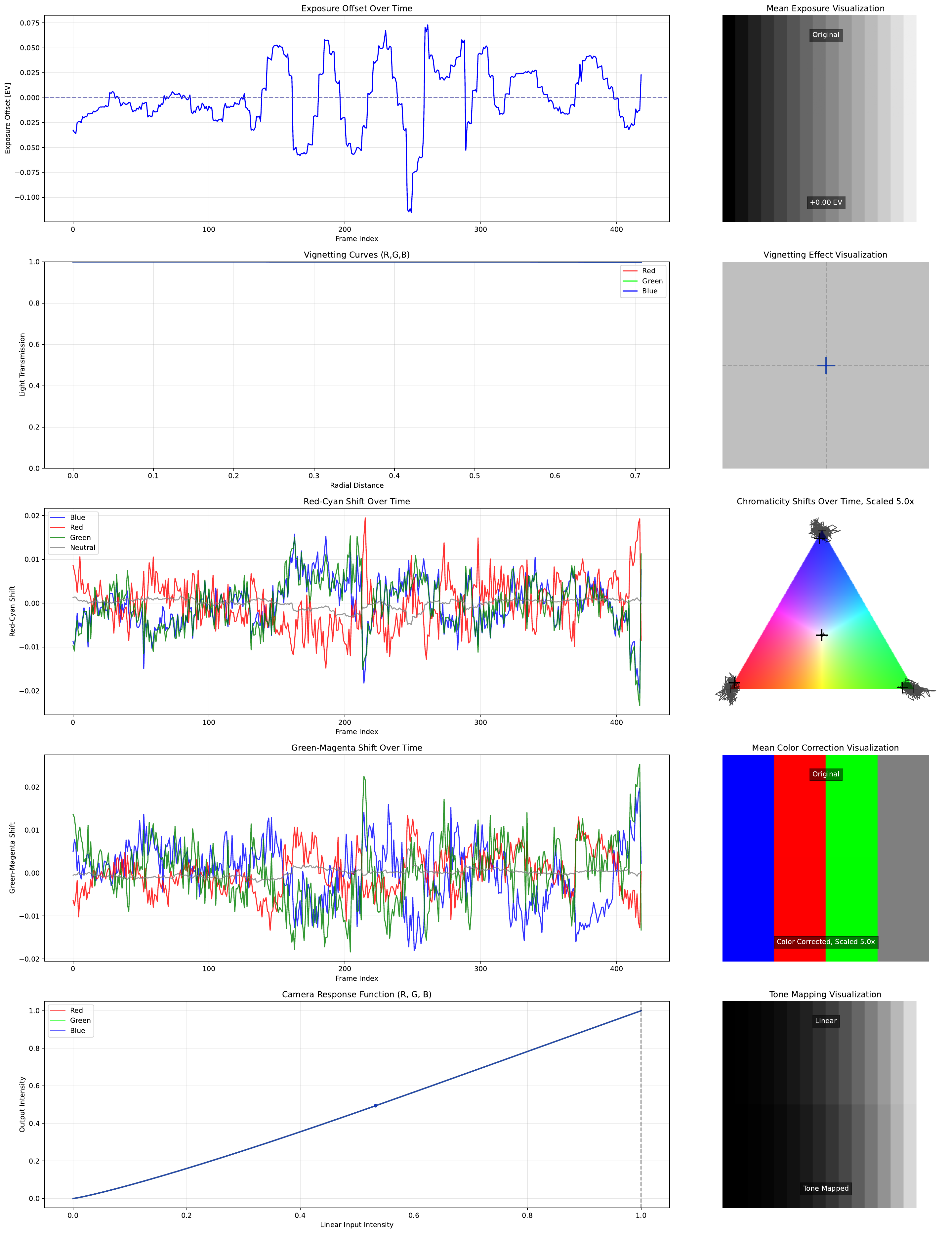}
    \caption{\ppisp{} evaluation report for the \adrasj{} dataset.}
    \label{fig:ppisp_report}
\end{figure}

We generated the \ppisp{} evaluation report by querying the trained module at the saved checkpoint using the evaluation script provided in the official repository \cite{deutsch2026ppispphysicallyplausiblecompensationcontrol}. As shown in Figure~\ref{fig:ppisp_report} for the \adrasj{} dataset, the module primarily learned to correct per-frame exposure variations, with the exposure offset ranging from approximately $-0.1$ to $+0.075$ EV across frames, while the vignetting and color corrections remained minimal. The camera response function was found to be approximately linear across all three color channels. In contrast, as shown in Figure~\ref{fig:ppisp_report_iss}, the STS-119 dataset exhibits significantly more aggressive corrections across all modules, with the exposure offset ranging from approximately $-0.48$ to $+0.49$ EV with sharp spikes, notable vignetting correction toward the image edges, and per-frame color corrections across all channels. The camera response function was found to be non-linear, suggesting a tone-mapped response that differs from the approximately linear response estimated for the \adrasj{} dataset.




\section{Discussion}
\label{sec:discussion}

In this section, we discuss the results obtained from each stage of the pipeline, highlighting the key findings, limitations, and potential directions for future work. Neural implicit surface reconstruction methods such as NeRF, \neuralangelo{}, and others require accurate camera poses as input rather than raw images, as the poses define the geometry of the scene. In controlled simulation environments, camera poses can be directly obtained from the simulator since the camera trajectory is known by design. However, in real-world scenarios such as on-orbit inspection, only the raw imagery is available and the camera poses must be estimated from the images themselves. This is where \sfm{} algorithms such as \colmap{} play a crucial role in recovering the camera intrinsics and per-frame poses directly from the inspection footage. As mentioned in Section~\ref{sec:Methodology}, \colmap{} initially failed to register the frames from both real datasets due to per-frame varying background, with only a few frames successfully registered across multiple experiments. After incorporating \samthree{} to isolate the object of interest from the background, \colmap{}'s performance improved significantly. As demonstrated in our prior work \cite{gopu2026dynamic}, the quality of \neuralangelo{} reconstruction is primarily governed by two factors: the quality of the camera poses and the photometric consistency of the input imagery across views. The low mean reprojection errors reported in Table~\ref{tab:colmap_results}, enabled by the background removal of \samthree{}, indicate accurate camera pose estimation, which in turn enabled \neuralangelo{} to reconstruct the 3D surface of both the ISS and the H-IIA upper stage. However, while accurate camera poses are necessary, they are not sufficient for high-fidelity reconstruction. The second factor, photometric consistency, refers to the consistency of brightness and color across views. This presents a significant challenge in on-orbit inspection scenarios, where illumination varies continuously, as reflected in the darker regions of the reconstructed mesh in Figure~\ref{fig:upperstage_mesh}.


To further validate this for the \adrasj{} dataset, we present a side-by-side comparison of the reconstructed meshes with and without \ppisp{} across four viewpoints in Figures~\ref{fig:ppisp_comparison_viewpoint_1}--\ref{fig:ppisp_comparison_viewpoint_4}. The geometry of the reconstructed meshes with and without \ppisp{} is identical, indicating that \ppisp{} did not affect the underlying geometry of the H-IIA upper stage. The difference lies in the texture brightness, as the fly-around maneuver captured imagery across both sunlit and shadowed regions, resulting in noticeably darker regions on the meshes on the left. \ppisp{} post-processing was able to correct the overall exposure difference across the mesh, and also recovered fine details above the nozzle area. However, both experiments struggled to recover the complete nozzle. The large cylindrical body section in the middle is the most consistently recovered feature across all viewpoints. 

Figures~\ref{fig:ppisp_comparison_viewpoint_2} and \ref{fig:ppisp_comparison_viewpoint_3} show the most visible improvement overall. \ppisp{} recovers a brighter tone across the cylindrical body that is severely underexposed in the mesh without \ppisp{}. However, Figure~\ref{fig:ppisp_comparison_viewpoint_3} still has some visible artifacts on the cylindrical body, which are likely artifacts from a hard cast shadow visible in the source imagery. \ppisp{} was only able to correct this partially. In contrast, in Figure~\ref{fig:ppisp_comparison_viewpoint_4}, the area between the cylindrical body and the nozzle presents the most notable improvement in this viewpoint. The mesh without \ppisp{} exhibits significant loss of detail in this region, whereas the mesh with \ppisp{} recovers these details alongside the brightness correction.

In contrast, the STS-119 dataset exhibits a different behavior. As shown in Figures~\ref{fig:iss_ppisp_comparison_viewpoint1}  and \ref{fig:iss_ppisp_comparison_viewpoint2}, \ppisp{} over-corrects on this dataset, introducing global darkening across the mesh in the first viewpoint and a strong warm color cast across the solar panels and the modules in the second viewpoint. However, Figures~\ref{fig:iss_ppisp_comparison_viewpoint3} and \ref{fig:iss_ppisp_comparison_viewpoint4} present a contrasting observation, where \ppisp{} reduces the overexposed regions visible in the baseline mesh, particularly on the white module surfaces, resulting in a more consistent appearance across the surface. The over-correction observed in the first two viewpoints may be due to the structural complexity of the ISS and the illumination conditions of the STS-119 footage, and remains an open question for future work.

Furthermore, the PSNR values remain similar across both experiments, with a mean of 35.46 dB for the baseline and 35.27 dB for \ppisp{} on the \adrasj{} dataset, and 32.36 dB for the baseline and 32.53 dB for \ppisp{} on the STS-119 dataset. The differences of less than 0.2 dB across both datasets suggest that PSNR alone does not reflect the qualitative differences observed in the reconstructed meshes.

Despite these promising results, several limitations remain. First, because no official ground truth CAD model of the H-IIA upper stage is publicly available, we are only able to perform a qualitative evaluation of the reconstructed mesh. Second, any segmentation errors from \samthree{} would propagate through the remaining stages of the pipeline. Third, both reconstructed meshes contain artifacts in the regions that were predominantly shadowed during the fly-around maneuver, as \neuralangelo{} assumes photometric consistency across frames. 

The results of this work motivate several directions for future investigation. First, improving the robustness of the \samthree{} segmentation would be beneficial, particularly in frames where the background introduces ambiguity, as this would reduce the risk of errors propagating through the remaining stages of the pipeline. Second, the illumination conditions inherent to on-orbit imagery present a fundamental challenge for neural implicit surface reconstruction methods that assume photometric consistency across views. Future work should focus on developing methods that explicitly account for these conditions to improve the fidelity of the reconstructed meshes. Third, evaluating the pipeline on additional on-orbit inspection footage with more challenging photometric conditions, such as lens flare and bloom, would deepen our understanding of \ppisp{}'s integration into \neuralangelo{}, potentially leading to improved reconstruction fidelity. Finally, a more comprehensive quantitative evaluation of the pipeline, including perceptual metrics, would better capture the improvements observed in the qualitative analysis.

\pagebreak{}
\section{Conclusion}
\label{sec:conclusion}

In this work, we present a pipeline for neural implicit surface reconstruction of non-cooperative resident space objects from real on-orbit inspection imagery, without ground truth camera poses or a reference CAD model. We demonstrate the pipeline on publicly released ISS inspection footage from the STS-119 mission and publicly released on-orbit inspection footage of an H-IIA rocket upper stage. Our results demonstrate that segmentation-based background removal via \samthree{} was critical to achieving full camera registration, and that photometric correction via \ppisp{} improved reconstruction fidelity, with the degree of improvement influenced by the illumination characteristics and structural complexity of the target. Using the footage captured during on-orbit missions, our pipeline produces a 3D mesh that can directly support debris characterization, structural assessment, active debris removal mission planning in simulation, and generation of synthetic datasets for computer vision research. Future work includes refining the \samthree{} segmentation strategy, developing illumination-aware reconstruction methods suited to on-orbit imagery, and evaluating our pipeline on additional inspection footage with more challenging photometric conditions.

\section*{Acknowledgments}
The authors would like to thank the United States Space Force and Air Force for supporting this research under Grant No. FA8750-24-C-B097.

\bibliographystyle{elsarticle-num}
\bibliography{references}

\end{document}